\documentclass{article}
\usepackage{microtype}
\usepackage{graphicx}
\usepackage{subfigure}
\usepackage{booktabs} 
\usepackage{mathtools}
\usepackage{bbm}
\usepackage{amsthm,amsmath,amssymb,amsfonts,bm}
\usepackage[top=2cm, bottom=2cm, left=2.5cm, right=2.5cm]{geometry}
\usepackage{hyperref}
\usepackage{wrapfig}

\usepackage{multicol}
\usepackage{multirow}
\usepackage{enumitem}
\usepackage{natbib}
\usepackage{url}
\usepackage{float}
\usepackage{xcolor} 
\usepackage{algorithm}
\usepackage[noend]{algpseudocode}
\usepackage{fancybox}
\cornersize{0.2} 
\usepackage{tcolorbox}
\usepackage[textsize=tiny]{todonotes}

\def\vf{{\bm{f}}}
\def\vg{{\bm{g}}}

\def\vx{{\bm{x}}}
\def\vy{{\bm{y}}}

\def\mW{{\bm{W}}}

\DeclareMathAlphabet{\mathsfit}{\encodingdefault}{\sfdefault}{m}{sl}
\SetMathAlphabet{\mathsfit}{bold}{\encodingdefault}{\sfdefault}{bx}{n}

\def\gA{{\mathcal{A}}}

\def\gD{{\mathcal{D}}}

\def\gO{{\mathcal{O}}}

\def\gX{{\mathcal{X}}}
\def\gY{{\mathcal{Y}}}

\def\sC{{\mathbb{C}}}

\def\sE{{\mathbb{E}}}

\def\sM{{\mathbb{M}}}

\def\sP{{\mathbb{P}}}

\def\sS{{\mathbb{S}}}

\newcommand{\R}{\mathbb{R}}

\DeclareMathOperator*{\argmax}{arg\,max}

\renewcommand{\tilde}{\widetilde}
\renewcommand{\hat}{\widehat}
\newcommand{\locs}{\mathrm{sgd}}

\newcommand{\glo}{\mathrm{agg}}

\usepackage[capitalize,noabbrev]{cleveref}
\theoremstyle{plain}
\newtheorem{theorem}{Theorem}[section]

\newtheorem{lemma}[theorem]{Lemma}

\theoremstyle{definition}

\theoremstyle{remark}

\newcommand{\eg}{{\em e.g.,~}}           
\newcommand{\ie}{{\em i.e.,~}} 
\newcommand{\ours}{{FedMT}}

\title{FedMT: Federated Learning with Mixed-type Labels}
\author{Qiong Zhang\\
Institute of Statistics and Big Data\\
Renmin University of China
\and
Jing Peng$^*$\\
Institute of Statistics and Big Data\\
Renmin University of China
\and
Xin Zhang\thanks{contributed equally}\\
MetaAI\\
\and
Aline Talhouk\\
Department of Obstetrics \& Gynecology\\
The University of British Columbia
\and 
Gang Niu\\
RIKEN
\and
Xiaoxiao Li\thanks{ Corresponding to: Xiaoxiao Li (xiaoxiao.li@ece.ubc.ca).}\\
Department of Electrical and Computer Engineering\\
The University of British Columbia
}

\begin{document}
\maketitle

\begin{abstract}
In \emph{federated learning}~(FL), classifiers (e.g., deep networks) are trained on datasets from multiple data centers without exchanging data across them, which improves the \emph{sample efficiency}.
However, the conventional FL setting assumes the \emph{same labeling criterion} in all data centers involved, thus limiting its practical utility. 
This limitation becomes particularly notable in domains like \emph{disease diagnosis}, where different clinical centers may adhere to different standards, making traditional FL methods unsuitable.
This paper addresses this important yet under-explored setting of FL, namely FL with \emph{mixed-type labels}, where the allowance of \emph{different labeling criteria} introduces inter-center label space differences.
To address this challenge effectively and efficiently, we introduce a model-agnostic approach called \ours{}, which estimates label space correspondences and projects classification scores to construct loss functions. 
The proposed \ours{} is versatile and integrates seamlessly with various FL methods, such as FedAvg.
Experimental results on benchmark and medical datasets highlight the substantial improvement in classification accuracy achieved by \ours{} in the presence of mixed-type labels.
\end{abstract}

\section{Introduction}
\label{sec:introduction}
Federated learning (FL) allows centers to collaboratively train a model while maintaining data locally, avoiding the data centralization constraints imposed by regulations such as the California Consumer Privacy Act~\citep{ccpa}, Health Insurance Portability and Accountability Act~\citep{act1996health}, and the General Data Protection Regulation~\citep{gdpr}.
This approach has gained popularity across various applications. 
Well-established FL methods, such as FedAvg~\citep{mcmahan2017communication}, employ iterative optimization algorithms for joint model training across centers. 
In each round, individual centers perform stochastic gradient descent (SGD) for several steps before communicating their current model weights to a central server for aggregation.

In the conventional FL classification framework, a classifier is trained jointly, assuming the same labeling criterion across all participating centers.
However, in real applications such as healthcare, the standards for disease diagnosis may be different across clinical centers due to the varying levels of expertise or technology available at different centers. 
Different centers may adhere to distinct diagnostic and statistical manuals~\citep{epstein2013changes,mckeown2015impact}, making it difficult to enforce a unified labeling criterion or perform relabeling, especially when labeling based on studies that cannot be replicated. Consequently, this results in disparate label spaces across centers. 
Furthermore, the center with the most intricate labeling criterion, crucial for future predictions thus referred as \emph{desired label space}, typically has a limited number of samples due to labeling complexity or associated costs. We consider this practical but underexplored scenario as learning from \emph{mixed-type labels} and aim to answer the following important question:
\begin{tcolorbox}[colback=white, colframe=black, arc=4pt, center title, fonttitle=\bfseries]
\begin{center}
{In the context of labeling criteria varying across centers, how can we effectively incorporate the commonly used FL pipeline (\eg FedAvg) to jointly learn an FL model in the desired label space?}
\end{center}
\end{tcolorbox}
\begin{figure*}[ht]
    \centering
    \includegraphics[width=0.9\textwidth]{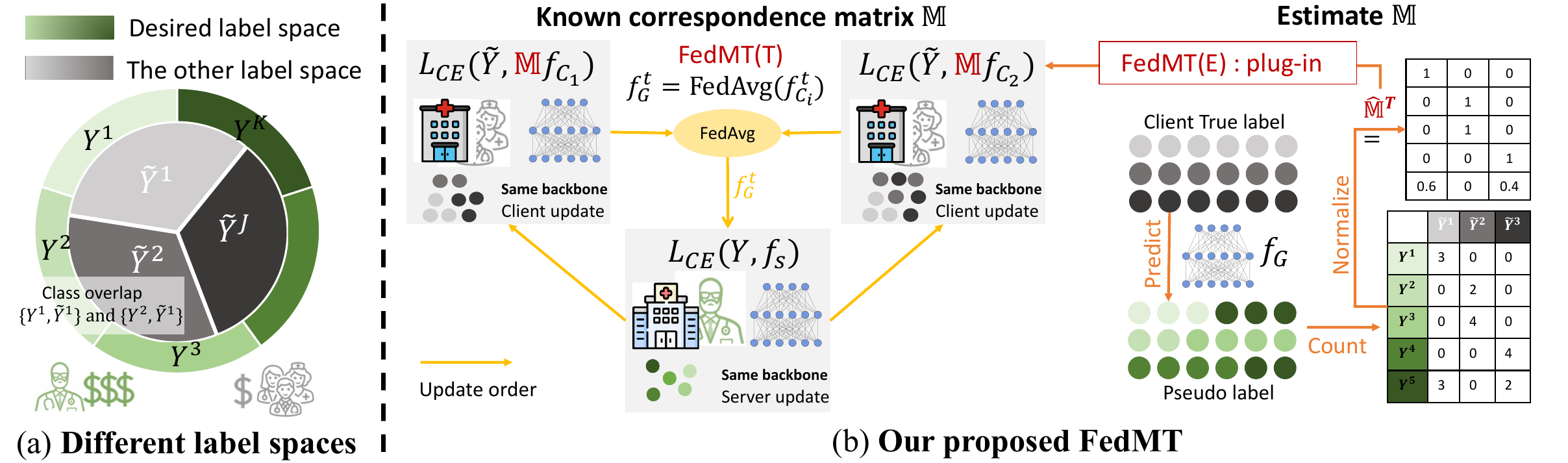}
    \caption{\small{\textbf{Illustration of the problem setting and our proposed \ours{} method.} (a) We consider different label spaces (\ie desired label space $\gY$ with $K$ classes and the other space $\tilde \gY$ with $J$ classes) where classes may overlap, such as $\tilde Y^1$ and $Y^2$. Annotation within the desired label space is typically more challenging and resource-intensive, resulting in a scarcity of labeled samples. (b) We employ a fixed label space correspondence matrix $\sM$ to establish associations between label spaces, effectively linking $\tilde \gY$ and $\gY$. Our method, denoted as \ours{} (T), locally corrects class scores $f$ using $\sM$ within the FedAvg framework. In instances where the correspondence matrix $\sM$ is unknown, we propose a pseudo-label based method to estimate $\hat\sM$. Subsequently, \ours{} (E) incorporates $\hat\sM$ into the loss function to correct class scores.}}
    \label{fig:main}
\end{figure*}
\noindent
\textbf{Problem setting:} 
To address the question, we consider a simplified but generalizable setting as illustrated in Fig.~\ref{fig:main} that two types of labeling criteria exist in FL. The label spaces are \emph{not necessarily nested}, namely, it is possible for a class in one label space to overlap with multiple classes in another space (\eg disease diagnoses often exhibit imperfect agreement). Additionally, drawing inspiration from a healthcare scenario, we assume a \emph{limited} availability of labeled data ($<5\%$) within the {desired label space}. 
One supercenter adheres to the complex labeling criterion according to the desired label space, while others use another distinct simpler labeling criterion.  The supercenter serves as the coordinating server for FL but also engages in local model updating like other clients. 
All centers jointly train an FL model following the standard FL training protocol, as shown in Fig.~\ref{fig:main} (b).

Under the problem setting described above, alternative approaches to handling different label spaces include personalized FL~\citep{collins2021exploiting}. 
However, these methods often neglect to exploit the inherent correspondence between different label spaces. 
Alternatively, transfer learning~\citep{yang2019federated}, which involves pre-training a model in one space and fine-tuning it in other spaces, can be an alternative solution in FL, but suboptimal pre-training may lead to negative transfer~\citep{chen2019catastrophic}. Other centralized strategies that require the pooling of all data features for similarity comparison through complex training strategies~\citep{hu2022weakly} increase privacy risk and are impractical for widely used FL methods like FedAvg.
In light of the limitations associated with these methods, we aim to address two key challenges: a) simultaneously leveraging different types of labels and their correspondences without additional feature exchange, and b) learning the FL model end-to-end. 

In this work, we introduce a plug-and-play method called \ours{}. This versatile strategy seamlessly integrates with various FL pipelines, such as FedAvg. 
Specifically, we use models with identical architectures, with the output dimension matching the number of classes in the desired label space across all centers. 
To utilize client data from other label spaces for supervision, we employ a probability projection to align the two spaces by mapping class scores.

\noindent 
\textbf{Contributions:} Our contributions are multifaceted. 
First, we explore a novel and underexplored problem setting--FL under mixed-type labels. 
This is particularly significant in real-world applications, notably within the realm of medical care.
Second, we propose a novel and versatile FL method, \ours{}, which is a computationally efficient and versatile solution; 
Third, we provide theoretical analysis on the generalization error of learning from data using mixed-type with projection matrix;
Lastly, our approach shows better performance in predicting in the desired label space compared to other methods, demonstrated by extensive experiments on benchmark datasets in different settings and real medical data. 
Additionally, we are also able to predict in the other space as a byproduct and we observe improved classification compared with other feasible baseline methods.

\section{Related Work}
\label{sec:relate}
\textbf{Federated learning (FL)}
FL is emerging as a learning paradigm for distributed clients that bypasses data sharing to train local models collaboratively. 
To aggregate the model parameters, FedAvg~\citep{mcmahan2017communication} would be the most widely used approach in FL. 
FedAvg variants have been proposed to improve optimization~\citep{reddi2020adaptive,rothchild2020fetchsgd} and for non-iid data~\citep{li2021fedbn,li2020federated,karimireddy2020scaffold}. 
Semi-supervised FL~\citep{jeong2020federated,bdair2021fedperl,diao2022semifl} considers scenarios where client samples are unlabeled. 
In contrast, our client data contains labeled samples but from different label spaces. 
Additionally, FL with samples from a single class on each client~\citep{yu2020federated} differs from our setting, as all samples are labeled using the same criterion in theirs.

\noindent
\textbf{Learning with labels of varying granularity}
In centralized setting, learning with labels of different granularity often involves the coarse-to-fine (C2F) label learning scenario, aiming to learn fine-grained labels given a set of coarse-grained labels. 
Notably, approaches such as~\citet{chen2021weak, touvron2021grafit, taherkhani2019weakly, ristin2015categories} heavily rely on the hierarchical assumption of coarse and fine labels, assuming knowledge of the hierarchical structure.
In contrast, our proposed method refrains from assuming the hierarchical structure of the two sets of labels, and we do not presume knowledge of this hierarchical structure. 
Moreover, centralized C2F methods, including~\citet{touvron2021grafit, hu2022weakly}, typically necessitate access to and communication of features from both types of labels (and consequently different clients) to calculate their losses. 
In our problem setting, each client possesses only one type of label. 
Extend these centralized C2F methods to the FL setting would incur additional communication costs and increase privacy risk.
\section{Methods}
\label{sec:setting}
In this section, we first overview the classical FL approach, then present the mathematical formulation of our problem and the proposed method.

\subsection{Problem Formulation}
We address a classification problem in which different labeling criteria are used at various data centers. 
For the sake of clarity, we consider two labeling criteria, denoted as $\gY= \{Y^k\}_{k=1}^K$ and $\tilde \gY= \{\tilde Y^j\}_{j=1}^J$ with $K>J$.

In this scenario, let $(\vx, y, \tilde y) \in \gX\times \gY\times \tilde\gY$, where both $y$ and $\tilde y$ serve as labels for the feature $\vx$ under two different labeling criteria. 
The key constraint is the observation of only one label at each data center. 
Driven by applications in the medical field, a `specialized center' is established, and its limited dataset originates from the desired label space. 
This center is referred to as a server, and its dataset is denoted as $\gD^{s}=\{(\vx_i^s, y_i^s): i\in [nK]\}$, where $y_i^s\in \gY$.
Furthermore, a total of $N$ labeled data, characterized by a different criterion, is distributed across $C$ clients. 
For each $c\in[C]$, $S_{c}$ denotes the indices of data in $\gD^{c} = \{(\vx_{i}^c, \tilde{y}_i^c): i\in S_c\}$ on the $c$-th client. Here, $\tilde{y}_i^c\in \tilde \gY$ represents labels from the alternative label space.
The corresponding unobserved label in the desired label space is denoted as $y_i^c$. 
Importantly, we assume disjoint datasets across all centers, \ie $S_i \cap S_j = \emptyset$. 
Let $N_c=|S_{c}|$, and the total labeled data is given by $N=\sum_{c\in[C]}N_{c}$.
Our objective is to train a global classifier in the desired label space denoted as $\vf: \gX\to\Delta_{K-1}$ using data from different label spaces within the system.

\subsection{Preliminary: Classical FL}
We begin by examining the classical Federated Learning (FL) approach, specifically FedAvg~\citep{mcmahan2017communication}, wherein all centers adhere to the \textbf{same} labeling criterion. Consider the feature $\vx$ and label $y$ with  joint density $p(\vx,y)$. 
A $K$-class classifier $\vf:\gX\to \Delta_{K-1}$ models the class score as $p(y=k|\vx)=f_k(\vx)$, where $f_{k}$ denotes the $k$th element of $\vf$. 
The label is predicted via $\hat{y} = \argmax_{k\in [K]} f_{k}(\vx)$.

In the classical FL setup, each client $c\in[C]$ possesses an independent identically distributed (IID) labeled training set $\gD_c = \{(\vx^c_i,y^c_i)\}_{i=1}^{N_c}$ of size $N_c$.
The objective of classical FL is to enable $C$ clients to jointly train a global classifier $\vf$ that generalizes well with respect to $p(\vx,y)$ without sharing their local data $\gD_c$. 
With access to data in all centers, the overall risk is 
$R(\vf)=C^{-1}\sum_{c=1}^C \hat R_c(\vf;\gD_{c})$ where
\begin{equation}
\label{eq:ce}    
\hat{R}_c(\vf;\gD_c)=\frac{1}{N_c}\sum_{i=1}^{N_c}\ell_{\text{CE}}(\vf(\vx^c_i),y^c_i),
\end{equation}
the \emph{cross-entropy} loss $\ell_{\text{CE}}(\vf^c(\vx),y)=-\sum\nolimits_{k=1}^K\mathbbm{1}(y=k)\log f_k^c(\vx)$, and $\mathbbm{1}(\cdot)$ is the indicator function.

To minimize the overall loss $R(\cdot)$ without data sharing, FedAvg involves alternating between a few local stochastic gradient updates on clients using client data, followed by an average update on the server.

However, in cases where labeling criteria differ across data centers, class scores across clients may not align. 
For instance, in a deep neural network, the number of neurons in the last layer may vary due to mismatches in the number of classes on different clients, leading to the failure of the server to average the model weights.
Therefore, we propose a novel approach to deal with \textbf{different} labeling criteria.

\subsection{Proposed Method}
\label{sec:proposed_method}
For classification problems, the standard CE loss, denoted as $\ell_{\rm CE}$, assumes that the dimension of class scores matches the number of classes. 
In our scenario, we can construct the server CE loss $\hat R_{s}(\vf; \gD_s)$ as in~\eqref{eq:ce} using server data.
However, the class scores $\vf(\vx^{c}) \in \Delta_{K-1}$, while $\tilde{y}^c\in \tilde{\gY}$ is a $J$-class label on clients. 
This mismatch prevents the direct use of the conventional CE loss on clients, prompting the question of how to evaluate risk on these clients.
To leverage the communication efficiency of FedAvg, we employ identical backbones on both the server and the clients. 
Constructing loss functions on clients becomes crucial for our classification problem with mixed-type labels. 
The core idea behind our proposed method is to align class scores and labels through projection. 
Further details of the method are provided below.

\textbf{Client loss construction} Each client sample has an unobserved label in the desired label space $\gY$.
For a given image $\vx^{c}$, our objective is to find the $K$-class scores $\{\sP(y^{c} = Y^{k}|\vx^{c})\}_{k=1}^{K}$.
According to the law of total probability:
\begin{align*}
\sP(\tilde{y}^{c} =\tilde{Y}^{j}|\vx^{c} )
=&\sum_{k} \sP(\tilde{y}^{c} = \tilde{Y}^{j} | y^{c} = Y^{k},\vx^{c})\sP(y^{c} = Y^{k}|\vx^{c} )\\
=&\sum_{k}\sP(\tilde{y}^{c} = \tilde{Y}^{j} | y^{c} = Y^{k})\sP(y^{c} = Y^{k}|\vx^{c} ).
\end{align*}
Here, the last equality is based on the assumption of instance independence in conditional probability. 
This equality allows $K$-class scores to be expressed linearly in terms of $J$-class scores. 
Let $\sM \in [0,1]^{J\times K}$ with
\begin{equation}
\label{eq:transition_matrix}
\sM_{jk} = \sP(\tilde{y}^{c} = \tilde{Y}^{j} | y^{c} = Y^{k}).
\end{equation}
When $\sM$ is known, we naturally obtain the following local loss based on projection class scores:
\begin{equation}
\label{eq:naive_client_loss}
\hat R_{c}^{p}(\vf;\gD_{c})
= \frac{1}{N_c}\sum_{i=1}^{N_c} \ell_{\text{CE}}(\sM \vf(\vx_i^{c}),\tilde{y}_i^{c})
\end{equation}
for any $c\in[C]$.
As each local loss involves only samples from the corresponding client and does not require coordination with other clients, we could jointly optimize all data centers using general FL strategies, such as FedAvg~\citep{mcmahan2017communication} we use, or other variants such as~\citet{li2020federated},~\citet{li2021fedbn}, and~\citet{karimireddy2020scaffold}.

Moreover, it is worth emphasizing that our proposed method readily extends to scenarios where label spaces differ across clients. 
This extension can be achieved by integrating the client-specific correspondence matrix, denoted as $\sM^{c}$, into the projection-based local loss~\eqref{eq:naive_client_loss}.

\subsection{Estimation of the correspondence matrix} 
The correspondence matrix $\sM$ plays a crucial role in our proposed method. 
While the method description above assumes knowledge of the projection matrix, practical application often involves direct computation of $\sM$ using domain knowledge. 
However, such domain knowledge is not available in some cases. 
To address this, we introduce an effective procedure for estimating $\sM$ from the data.

\begin{algorithm}[!t]
    \caption{FL using \ours{} (Ours)}
    \label{alg:2}
    \textbf{Server Input:}  server model $\vf^s$, \textit{small} server dataset $\gD_s=\{(\vx^s,\vy^s)\}$ where $\vy^s\in\gY$
    \\
    \textbf{Client Input:}
    aggregation step-size $\eta_{\rm agg}$,
    and global communication round $R$, regularization parameters $\lambda_1$, $\lambda_2$, client dataset $\gD_c=\{(\vx^c,\vy^c)\}$ 
    \begin{algorithmic}[1]
    \State {For $r = 1 \to R$ rounds, run \textbf{A} on server and \textbf{B \&C} on each client iteratively.}
    \Procedure{\textbf{A}.ServerModelUpdate}{$r$}
    \State{$\vf^s\leftarrow\vf$} \Comment{Receive updated model from \textsc{proc. C}}
    \For{$\tau=1\to t$}
    \State{$\vf^s\leftarrow\vf^s-\eta_{\locs}\cdot\nabla\hat{R}_s(\vf^s;\gD_s)$}
    \EndFor
    \State{send $\vf^s$ to \textsc{proc. B}}
    \EndProcedure{}
    \Procedure{\textbf{B}.ClientModelUpdate}{$r$}
    \State{$\vf^c\leftarrow\vf$} \Comment{Receive updated model from \textsc{proc. A}}    
    \State{Generate pseudo-labels $\hat{y}_{i}^{c} = \vf^{c}(\vx_{i}^{c})$}
    \State{Construct high-confidence sample subset 
    $S_{c}^{\text{fix}}(\xi) = \{i\in S_{c}: \max_{k} f_k^{c}(\vx_i^{c}) > \xi\}$}
    \If{$S_{c}^{\text{fix}}(\xi) = \emptyset$} 
    \State{stop \& return.}
    \Else
    \State{Construct an equal-size mixup subset
    $S_{c}^{\text{mix}}(\xi) = \text{Sample } |S_{c}^{\text{fix}}(\xi)| \text{ with replacement from } S_c\}$}
    \EndIf
    \For{$\tau=1\to t$}
    \For{batch $(x^{\textup{fix}}_{b}, \hat y^{\textup{fix}}_{b}), (x^{\textup{mix}}_{b}, \hat y^{\textup{mix}}_{b})$ with $b\in S_c^{\textup{fix}}, S_c^{\textup{mix}}$}
    \State{$\lambda_{\text{mix}} \sim \text{Beta}(a, a)$}
    \State{$x_{\text{mix}} \leftarrow \lambda_{\text{mix}} x^{\text{fix}}_{b} + (1-\lambda_{\text{mix}}) x^{\text{mix}}_{b}$}
    \State{$L_{\text{fix}} \leftarrow \ell(\vf(\gA(x^{\text{fix}}_{b})), \hat y^{\text{fix}}_{b})$}
    \State{$L_{\text{mix}} \leftarrow \lambda_{\text{mix}}\ell(\vf(x_{\text{mix}}), \hat y^{\text{fix}}_{b}) + (1-\lambda_{\text{mix}})\ell(\vf(x_{\text{mix}}), \hat y^{\text{mix}}_{b}))$}
    \If{$\sM$ is not known}
    \State{Estimate $\hat\sM$ via~\eqref{eq:m_hat}}
    \EndIf
    \State {Construct local loss $\hat R_{c}^{\text{reg}}(\vf^c;\gD_c)$ via~\eqref{eq:strong_augmented_local_loss}} 
    \State{$\vf^{c} \leftarrow \vf^{c}-\eta_{\locs} \nabla \hat R_{c}^{\text{reg}}(\vf^c;\gD_c)$}
    \EndFor
    \EndFor
    \State{send $\vf^l-\vf$ to \textsc{proc. C} for $l \in [C]$}
    \EndProcedure{}
    \Procedure{\textbf{C}. ModelAgg}{$r$} 
        \State{receive model updates from \textsc{proc. B}}
        \State{$\vf\leftarrow\vf-\eta_{\glo}\cdot\sum\nolimits_{c \in \{[C]\}}(\vf^c-\vf)$}
        \State{broadcast $\vf$ to \textsc{proc. A}}
    \EndProcedure{}
    \end{algorithmic}
\end{algorithm}

Recalling the definition of the correspondence matrix in~\eqref{eq:transition_matrix}, when both sets of labels are known for each sample, a practical estimate for $\sM_{jk}$ is to use its empirical version $\tilde{\sM}_{jk} = \sum_{i\in S_{c}} \mathbbm{1}(\tilde{y}_i^{c} = \tilde{Y}^{j}, y_i^{c}= Y^k)/\sum_{i\in S_{c}} \mathbbm{1}(y_i^{c} = Y^k)$.
However, a challenge arises with this estimate as the true label $y_i^{c}$ in the desired label space remains unknown for client data. 
To overcome this, we use the aggregated model to predict $y_i^{c}$ and view the predicted pseudo-label as the true label for estimation. 
To enhance the efficiency of the estimator, we estimate the correspondence matrix using only samples with high-confidence pseudo-labels.

In particular, let $\vf^{r}$ be the aggregated model weight in the $r$th communication round.
Let $\hat{y}_i^{c} = \argmax_{k} f_k^{r}(\vx_i^{c})$ represent the pseudo-label for the $i$th sample on the $c$th client. 
Additionally, define
\[S_{c}^{\text{fix}}(\xi) = \{i\in S_{c}: \max_{k} f_k^{r}(\vx_i^{c}) > \xi\}\]
as the set of high-confidence samples on the $c$th client in the $r$th communication round. 
The confidence threshold $0 < \xi < 1$ serves as a preselected hyperparameter for all clients. 
If, for a specific client $c$, we find that $S_{c}^{\text{fix}}(\xi) = \emptyset$, then the process halts and the client refrains from transmitting data to the server. 
Otherwise, the correspondence matrix on $c$th client in $r$th communication round is defined as
\begin{equation}
\label{eq:m_hat}
\hat{\sM}_{jk}^{c} = \frac{\sum_{i\in S_{c}^{\text{fix}}(\xi)} \mathbbm{1}(\tilde{y}_i^{c} = \tilde{Y}^{j}, \hat y_i^{c}= Y^k)}{\sum_{i\in S_{c}^{\text{fix}}(\xi)} \mathbbm{1}(\hat y_i^{c} = Y^k)}.
\end{equation}
The estimated correspondence matrix also applies to the scenario where label spaces differ across clients.

\noindent
\textbf{Strong data augmentation}
To further improve the performance in our proposed method, we also use the technique of strong data augmentation inspired by~\citet{diao2022semifl}.
Specifically, we apply strong data augmentation loss as a regularization technique.
We generate a mixup dataset:
\[S_{c}^{\text{mix}}(\xi) = \{\text{Sample } |S_{c}^{\text{fix}}(\xi)| \text{ with replacement from } S_c\}\]
where $|S_{c}^{\text{fix}}(\xi)|$ denotes the number of elements of $S_{c}^{\text{fix}}(\xi)$. 
For each local training epoch of the client $c$, it randomly divides local data $S_{c}^{\text{fix}}(\xi), S_{c}^{\text{mix}}(\xi) $ into batches.
For each batch iteration, client $c$ constructs Mixup data from one particular data batch $(x^{\text{fix}}_{b}, \hat{y}^{\text{fix}}_{b}), (x^{\text{mix}}_{b},\hat{y}^{\text{mix}}_{b})$ by
\begin{equation*}
\lambda_{\text{mix}} \sim \text{Beta}(a, a), \quad x_{\text{mix}} \leftarrow \lambda_{\text{mix}} x^{\text{fix}}_{b} + (1-\lambda_{\text{mix}}) x^{\text{mix}}_{b}
\end{equation*}
where $a$ is the hyperparameter of Mixup~\citep{zhang2018mixup}.
Then the ``fix'' loss $L_{\text{fix}}$~\citep{sohn2020fixmatch} and the ``mix'' loss $L_{\text{mix}}$~\citep{berthelot2019remixmatch} are locally defined by
\begin{equation*}
\begin{split}
L_{\text{fix}} =& \ell(\vf(\mathcal{A}(x^{\text{fix}}_{b}), \hat y^{\text{fix}}_{b})\\
L_{\text{mix}} =& \lambda_{\text{mix}} \ell( \vf(x_{\text{mix}}), \hat y^{\text{fix}}_{b}) + (1-\lambda_{\text{mix}})\ell(\vf(x_{\text{mix}}), \hat y^{\text{mix}}_{b})).     
\end{split}
\end{equation*}
Here, $\mathcal{A}$ represents a strong data augmentation mapping such as the RandAugment~\cite{cubuk2020randaugment}.  
Then we regularize each client loss with the strong data augmentation loss as follows:
\begin{equation}
\label{eq:strong_augmented_local_loss}
\hat R_{c}^{\text{reg}}(\vf; \gD_c) = \hat R_{c}^{p}(\vf; \gD_c) + \lambda_2(L_{\text{fix}} + \lambda_1 L_{\text{mix}})
\end{equation}
where
\[
\hat R_{c}^{p}(\vf; \gD_c)\label{eq:strong_augmented_local_loss}
= 
\begin{cases}
        \sum_{i\in S_{c}^{\text{fix}}(\xi)} \ell_{\text{CE}}(\sM \vf(\vx_i^{c}),\tilde{y}_i^{c})   & \text{if $\sM$ known,}\\
        \sum_{i\in S_{c}^{\text{fix}}(\xi)} \ell_{\text{CE}}(\hat\sM^c \vf(\vx_i^{c}),\tilde{y}_i^{c}) & \text{otherwise.}
\end{cases}
\]
Our algorithm follows an iterative process, with each round comprising updates on the server followed by communication to each client for local adjustments. 
The weights after clients' updates are subsequently averaged using FedAvg. 
The detailed \ours{} algorithm is described in Algorithm~\ref{alg:2}.

\subsection{Generalization Error for \ours{}}
In this section, our focus centers on examining the generalization error of \ours{}. 
Instead of studying \ours{} in full generality, we restrict our attention to investigate how samples originating from two distinct label spaces contribute to improving training within both spaces. 
For a more thorough understanding, additional background information and technical details can be found in the Appendix~\ref{sec:appendix_theory}. 
We provide a simplified statement as follows.

\begin{theorem}[Informal]
\label{thm:generalization_error}
Let $\gD_{s} = {(\vx_i^s, y_i^s): i=1,\ldots, n}$ and $\gD_{c} = {(\vx_i^c, \tilde y_i^c): i=1,\ldots, m}$ be two datasets, and $\vg: \gX \to \Delta_{K-1}$ be any classifier based on the combined dataset $\gD_{s} \cup \gD_{c}$.
Let $\vf$ be same as before, define $\tilde{\vf}: \gX \to \Delta_{J-1}$ with its $j$th element as $\tilde{f}_j(\vx) = \sP(\tilde y = \tilde{Y}^{j} | X=\vx)$ 
and $\sS: \Delta_{J-1}\to \Delta_{K-1}$ where $\sS_{kj} = \sP(y = Y^{k}|\tilde{y} = \tilde{Y}^{j})$.
With a probability of at least $1-2\max(n,m)^{-1/2}$, we have
\begin{itemize}
    \item When $\sM$ is known, $\sE\|\vg(\vx)-\vf(\vx)\|_2 \leq$ empirical loss $+ \tilde{\gO}((m+n)^{-1/2})$.
    \item When $\sM$ is estimated via~\eqref{eq:m_hat}, $\sE\|\vg(\vx)-\vf(\vx)\|_2 \leq$ empirical loss $+ \tilde{\gO}((m+n)^{-1/2})+ \tilde{O}(m^{-1/2})$.
    \item The generalization error in two label spaces satisfies $\sE\|\sM\vg(\vx)-\tilde\vf(\vx)\|_2 \leq  \sqrt{K} \sE\|\vg(\vx)-\vf(\vx)\|_2.$
\end{itemize}
\end{theorem}
The first two generalization errors are computed for predictions within the desired label space, and the final one targets predictions within the other label space.
In the first bound, the error term scales in the order of $\tilde{\gO}((m+n)^{-1/2})$. 
This contrasts with a single dataset scenarios where the error terms, which improves from the situations with only one dataset where the errors are about $\tilde{\gO}(n^{-1/2})$ or $\tilde{\gO}(m^{-1/2})$. 
In the second bound, the estimation of the correspondence matrix introduces an additional error of order $\tilde{O}(m^{-1/2})$.
In the last bound, a notable point is that, as a byproduct result of our approach, the prediction error in the other label space has the same order as in the desired label space.
\section{Experiments}
\label{sec:experiment}
In this section, we demonstrate the effectiveness of our proposed method, \ours{}, when data labels are from different spaces. 
Compared to other training strategies and prior art,~\ours{} consistently achieves better test accuracy in predicting labels in the desired label space, as demonstrated in CIFAR100~\citep{krizhevsky2009learning} and a dermatological dataset called Fitzpatrick 17k~\citep{groh2021evaluating}.

\subsection{Benchmark Experiments Setup}
\label{exp:benchmark}

\textbf{Dataset and setting} 
We use the CIFAR100~\citep{krizhevsky2009learning} dataset to mimic our proposed problem setting, as it naturally provides two labels with different degrees of granularity. 
The CIFAR100 dataset consists of 50K images, associated with $K=100$ subclasses that could be further grouped into $J=20$ superclasses. 
The sub-class space is viewed as our desired label space and the super-class space as the other label space.
Based on the superclass information in CIFAR100, we are able to formulate the correspondence matrix $\sM$ for all clients, which is given in the Appendix~\ref{sec:detail_benchmark}.
We have $C+1$ data centers in total and one of the centers serves as the server to coordinate FL training.
The server has a small amount of data with sub-class annotations, and other centers are labeled using super-class annotations. 
Our objective is to train a classification model using FL to predict subclass labels using all data centers simultaneously.

To study the effect of the number of training samples in the desired label space on model performance, we randomly select $n$ observations from each of the $100$ subclasses on the server.
Unless otherwise specified, the rest of the observations in the training set are split into $C=10$ subsets completely at random, ensuring that each class is equally represented in each subset. 
Each subset corresponds to a dataset stored on one client, and we use $N_c=4000$ unless otherwise specified. 
We use the accuracy of the $K$ class prediction in the hold-out test set for performance comparison.
We list other training details, such as learning rate and optimizer, in Appendix~\ref{sec:detail_benchmark}.

\noindent
\textbf{Baselines}
We present a comprehensive comparison of the following approaches:
1) \textbf{Single}: This approach utilizes only the data from the desired label space to train the classifier;
2) \textbf{FedTrans}~\citep{chen2021meta}: All clients train a $J$-class classifier using FedAvg, which is subsequently fine-tuned on the server with a new $K$-class linear layer;
3) \textbf{FedRep}~\citep{collins2021exploiting}: In this approach, clients engage in training a $J$-class classifier, while the server concurrently trains a $K$-class classifier with a unique last layer possessing a distinct output dimension. The sharing of identical backbone parameters across classifiers is facilitated through the FedAvg algorithm. This design accommodates diverse last layers across different machines. Subsequently, the performance evaluation is conducted using the classifier trained on the server;
4) \textbf{SemiFL}~\citep{diao2022semifl}: This method treats client samples as unlabeled data, incorporates pseudo-labeling on the clients, and augments samples in supervised loss training. SemiFL stands as the state-of-the-art federated semi-supervised learning method.
Our proposed method is denoted as \ours{} (T) with a known correspondence matrix $\sM$, while \ours{} (E) refers to our proposed method with an estimated correspondence matrix $\hat\sM$.

To highlight the distinctions between the baselines and our proposed method, we provide an overview of the learning strategies employed by all methods in Table~\ref{tab:coceptual_comparison}.
\begin{table}[ht]
\caption{\small \textbf{Conceptual comparison of the baselines with \ours{}} in terms of the learning strategy and the type of labeled information used on the server and clients during the FL. }
\label{tab:coceptual_comparison}
\centering
\begin{tabular}{l| c c c c c}
\toprule
&Single&FedRep&FedTrans&SemiFL & \ours{}(Our)\\ 
\midrule
Strategy & Centralized & FL & FL+FT & FL & FL \\ 
Server & --  & $\gY$  & $\gY$ & $\gY$ & $\gY$ \\ 
Clients & -- & $\tilde{\gY}$& $\tilde{\gY}$ & pseudo & $\tilde{\gY}$\\ 
\bottomrule
\end{tabular}
\end{table}
The Single approach is centralized and only uses the server data from the desired label space.
In the FedTrans approach, clients first jointly learn a model in a FL fashion in the other label space and then fine tune (FT) the model on the server in the desired label space.
The rest of the approaches are FL in an end-to-end fashion.
For these FL-based methods, we list the type of labeled information used during training.
Recall that $\gY$ represents the desired label space and $\tilde{\gY}$ represents the other label space.

\subsection{Benchmark Experiments Results}
We present the main experiment results, additional results are given in Appendix~\ref{sec:appendix_more}.

\noindent
\textbf{Varying number of samples in the desired label space} 
We fix the number of clients ($C$) at 10 and vary the number of samples per class, represented by $n$, on the server across the range of ${5, 10, 15, 20, 25}$. 
In this experiment, we employ ResNet18~\citep{he2016deep} as the backbone architecture.

For subclass prediction, the Single, FedRep, and FedTrans methods, we observed superior performance when utilizing ImageNet pre-trained weight initialization compared to random initialization. 
The detailed comparison of the experimental results, with respect to different initialization strategies, can be found in the Appendix~\ref{app:initializations}.
On the other hand, for SemiFL and our proposed \ours{} approach, we discovered that they exhibit better performance with random initialization. 
Consequently, we provide the experimental results showcasing the preferred initialization strategy for each of these approaches in all our experiments conducted on CIFAR100.
\begin{table}[ht]
\centering
\caption{\small \textbf{Comparison with baselines.} Comparison of the test accuracy for 100-class classification using our methods and alternative methods on the CIFAR100 benchmark dataset with different per sub-class number $n$ images on the server. We report mean{\tiny{$\pm$sd}} from three trial runs. The best method is highlighted in boldface.}
\label{tab:cifar_varyn_noisefree}
\begin{tabular}{lcccccc}
\toprule
$n$& Single &  FedTrans & FedRep & SemiFL & \multicolumn{2}{c}{\textit{Ours}: \ours{} (T \& E)} \\ 
\midrule
5 & 15.80\scriptsize{$\pm$0.66} & 39.45\scriptsize{$\pm$0.49} & 43.43\scriptsize{$\pm$0.67} & 27.93\scriptsize{$\pm$0.37} & {\bf 57.12}\scriptsize{$\pm$1.12}& 46.41\scriptsize{$\pm$0.78}\\
10 & 20.38\scriptsize{$\pm$0.28}  &  42.97\scriptsize{$\pm$0.15} & 48.70\scriptsize{$\pm$0.29} & 40.80\scriptsize{$\pm$1.12} & {\bf 61.57}\scriptsize{$\pm$0.20}& 57.11\scriptsize{$\pm$0.13} \\
15& 27.16\scriptsize{$\pm$0.68} & 45.01\scriptsize{$\pm$0.26} & 50.49\scriptsize{$\pm$0.55} & 49.01\scriptsize{$\pm$0.57} &{\bf 64.55}\scriptsize{$\pm$0.95}&61.41\scriptsize{$\pm$0.56} \\
20 & 29.12\scriptsize{$\pm$0.80} & 46.57\scriptsize{$\pm$0.19} & 53.14\scriptsize{$\pm$0.34} & 53.42\scriptsize{$\pm$0.44}& {\bf 66.16}\scriptsize{$\pm$0.06}&64.83\scriptsize{$\pm$0.07}\\
25 & 29.57\scriptsize{$\pm$0.29} & 47.37\scriptsize{$\pm$0.20} & 54.83\scriptsize{$\pm$0.30} &55.98\scriptsize{$\pm$0.61}&{\bf 66.70}\scriptsize{$\pm$0.21}&66.01\scriptsize{$\pm$0.13} \\
\bottomrule
\end{tabular}
\end{table}
As shown in Tab.~\ref{tab:cifar_varyn_noisefree}, when $n$ increases (\ie the supervision information in the desired label space increases), the accuracy of all approaches improve. 
A comparison between FL-based methods and Single reveals that FL enhances model performance by accessing data from other label spaces.
Our proposed \ours{} (T \& E) consistently outperforms alternative methods across all scenarios, showcasing a synergistic collaborative effect with clients' data from other label spaces. 
Specifically, when the true correspondence matrix is known (\ours{} (T)), our method exhibits the highest performance. 
Even in cases where the true correspondence matrix is unknown and estimated from the data (\ours{} (E)), our approach maintains a performance comparable to that of situations where the true correspondence matrix is known, surpassing the performance of other methods.
Furthermore, our proposed method exhibits great performance even when the supervision information in the desired label space is limited (\ie $n$ is small), contrasting with the observation that the effectiveness of other methods, such as SemiFL, becomes more pronounced only with a larger number of labeled data in the desired label space.

As an additional advantage, our method inherently allows for predictions in the other label space. 
Therefore, we present the superclass test accuracy of both feasible baselines and \ours{}. 
We skip single and semiFL because they cannot predict coarse labels. 
The results are shown in Figure~\ref{fig:cifar_coarse}. 
Further information regarding the experimental setup is provided in Appendix~\ref{sec:coarse_prediction}.
As shown in the figure, FedRep and our method perform better as $n$ increases; this is due to the increase in the total sample size and hence the amount of information contained.
The FedTrans approach is tested using the superclass classifier trained using client data, and therefore does not leverage server information and remains constant regardless of the value of $n$.
Overall, our approach has better performance for superclass prediction compared to other baselines.

\begin{figure}[htpb]
\centering
\subfigure[Super-class predict.]{
\begin{minipage}[t]{0.48\columnwidth}
\label{fig:cifar_coarse}
\centering
\includegraphics[width=0.95\textwidth]{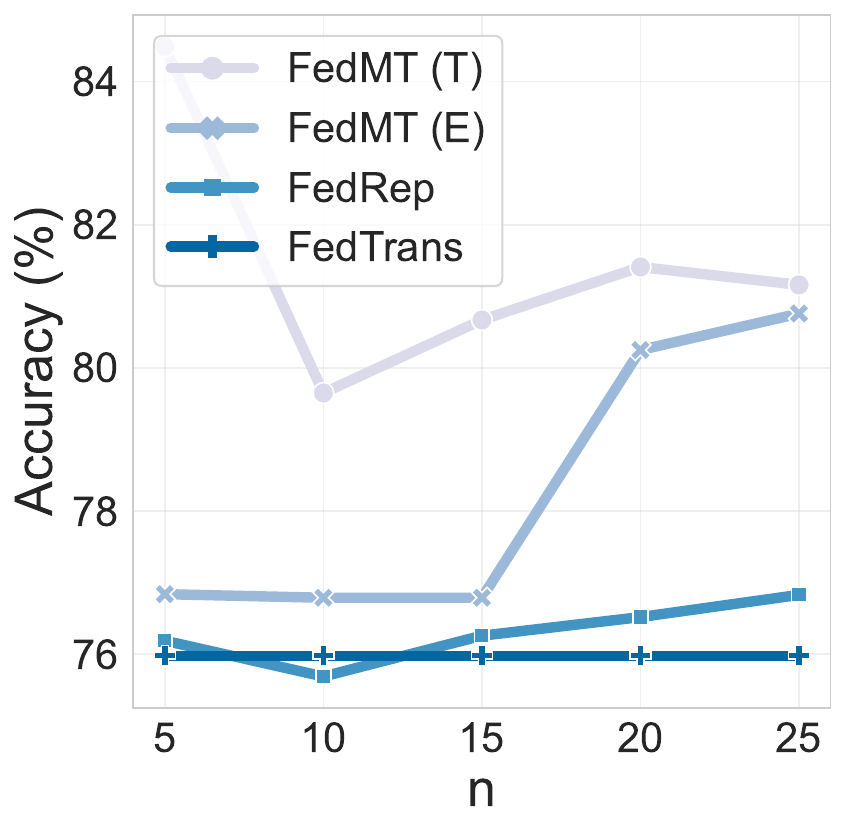}
\end{minipage}%
}%
\subfigure[IID vs Non-IID]{
\begin{minipage}[t]{0.48\columnwidth}
\label{fig:cifar_noniid}
\centering
\includegraphics[width=0.95\textwidth]{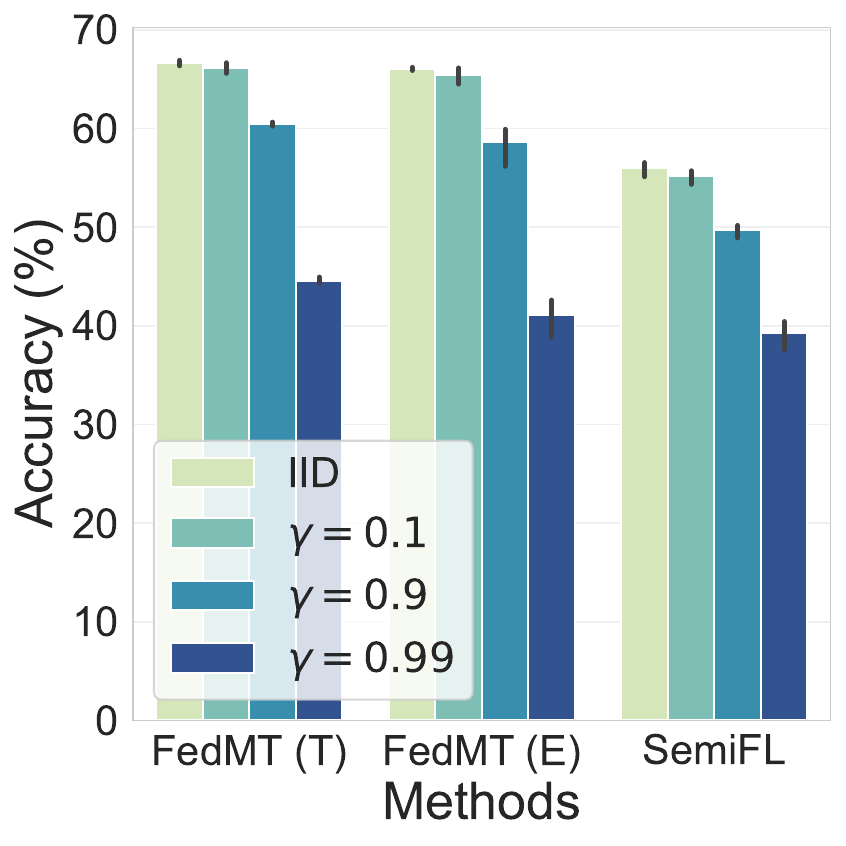}
\end{minipage}%
}\\%
\subfigure[Effect of client num.]{
\begin{minipage}[t]{0.48\columnwidth}
\label{fig:cifar_scalability}
\centering
\includegraphics[width=0.95\textwidth]{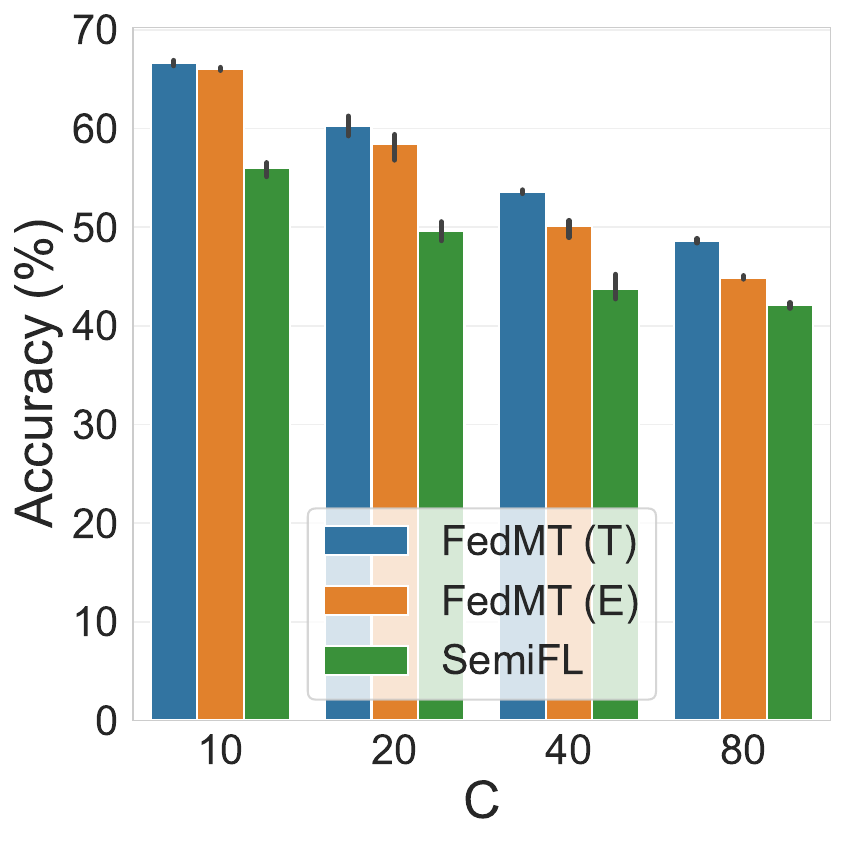}
\end{minipage}%
}
\subfigure[Effect of backbones]{
\begin{minipage}[t]{0.48\columnwidth}
\label{fig:cifar_backbone}
\centering
\includegraphics[width=0.95\textwidth]{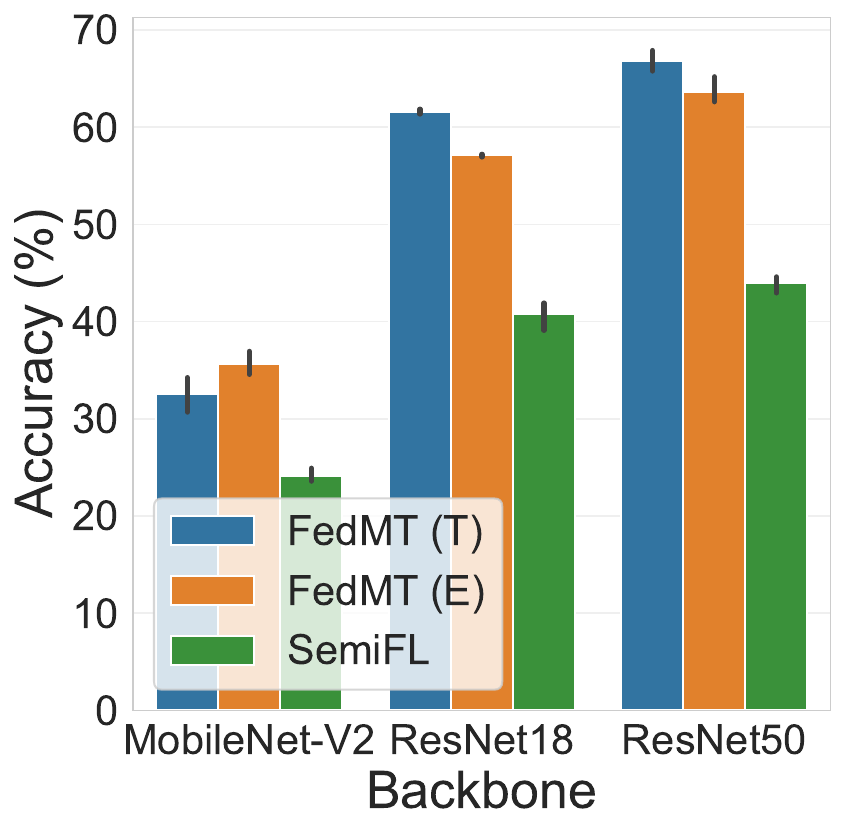}
\end{minipage}%
}
\caption{\small \textbf{Empirical analysis of \ours{} under various settings.}
Comparison with baselines is presented under different scenarios: (a) super-class prediction accuracy with varying numbers of sub-class samples,
 (b) in a non-iid setting, (c) with varying numbers of clients, and (d) with diverse backbone architectures.
}
\label{fig:cifar_comparison_more}
\end{figure}

\noindent
\textbf{Non-IID setting}
The heterogeneity of training data among clients poses a significant challenge in FL~\citep{kairouz2021advances}. 
To further investigate the robustness of our \ours{} method, we extend our analysis to the non-IID (non-identically distributed) FL setting, where clients may have different label distributions.
In the non-IID setting, each client has two leading majority classes, with most of the samples being from these two classes while the rest of the classes are the minority classes.
The details of the non-IID splitting of the dataset are given in Appendix~\ref{sec:non-iid-split}.
A higher value of $\gamma$ indicates a stronger level of label heterogeneity.
We evaluate the performance of both our proposed method and the leading baseline approach SemiFL and the experiment results are given in Fig.~\ref{fig:cifar_noniid}. 
It is evident from the figure that the performance of both approaches experiences a decline with an increase of label heterogeneity.
Nevertheless, our method consistently outperforms the leading baseline, SemiFL, even in the most extreme case ($\gamma=0.99$).

\noindent
\textbf{Client number scalability}
The scalability of federated learning, with regard to the number of participating machines, constitutes a critical factor influencing the efficiency and effectiveness of collaborative training processes. 
As such, we evaluate the performance of both our proposed method and the leading baseline approach, SemiFL, across varying numbers of clients. 
In this experimental setup, we maintain a consistent sample size of $n=25$ for all experiments while fixing the total number of samples across clients (i.e., $N_{c}\times C$) at $40$K.
For instance, with $N_c=2000$ when $C=20$. 
The experimental results are summarized in Fig.~\ref{fig:cifar_scalability}.
It is evident that the performance of both approaches experiences a decline with an increase in the number of clients. 
However, our proposed method, \ours{}, exhibits greater robustness to changes in the number of clients. The drop in SemiFL performance from $C=10$ to $C=80$ is $53.38\%$ (calculated as $(55.98-26.10)/55.98$), while \ours{} (E) demonstrates a comparatively milder decrease of only $35.59\%$.
Furthermore, our \ours{} exceeds the leading baseline regardless of the number of clients $C$.

\noindent
\textbf{Experiment result with different backbones}
We also test the performance of all approaches with different model architectures, namely MobileNet-V2~\citep{sandler2018mobilenetv2}, ResNet18, and ResNet50.
We initialize all methods and all backbones with pretrained weights on ImageNet.
The number of parameters increases from MobileNet-V2 to ResNet50.
Regardless of the architecture, our method has the best performance.

\begin{figure}[htbp]
\centering
\includegraphics[width=0.7\columnwidth]{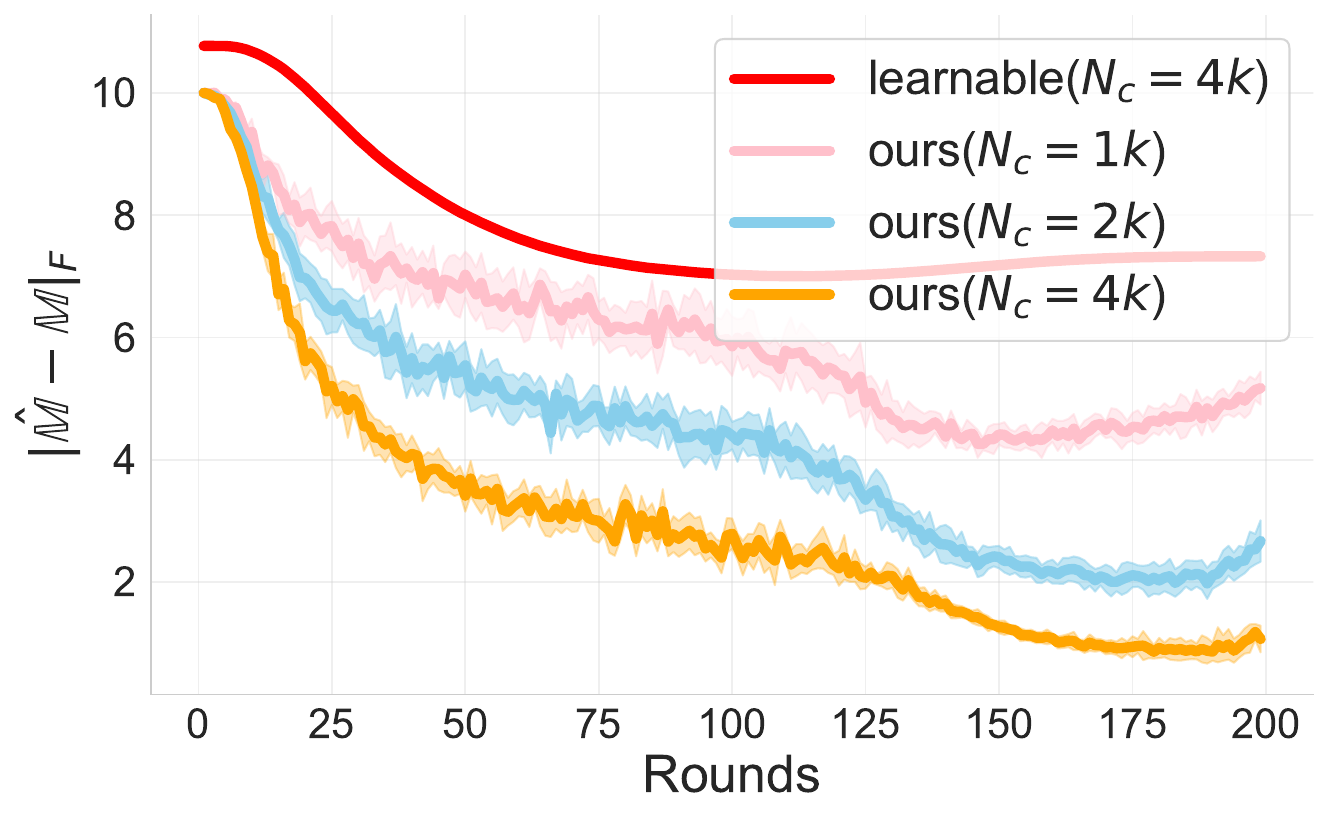}
\caption{\small \textbf{Estimation error of $\sM$.} The comparison of the Frobenius norm between $\hat\sM$ and true $\sM$ via two approaches (\ie \ours{} and learnable) and different client sample sizes ($N_c$).}
\label{fig:cifar_goodness_of_hatm}
\end{figure}

\noindent
\textbf{Goodness of the estimated correspondence matrix $\hat{\sM}$}
We propose to estimate the correspondence matrix via~\eqref{eq:m_hat}, and we empirically demonstrate that the accuracy of the estimator improves progressively during training. 
Specifically, we assess the Frobenius norm $\|\hat\sM^{c} - \sM\|_F$ for each client at every communication round. 
We visualize the mean and error bars of the Frobenius norm across $C=10$ clients as a function of communication rounds in Fig.~\ref{fig:cifar_goodness_of_hatm}. 

The figure illustrates that
a) the discrepancy decreases as the sample size of the client increases, indicating the consistency of our estimation method;
b) the discrepancy approaches zero as the number of rounds increases for large $N_c=4000$, indicating the efficacy of our estimation method.
Furthermore, we conduct a comparative analysis by contrasting our estimation method with an alternative approach in which $\sM$ is designated as a learnable correspondence matrix. 
Refer to Appendix~\ref{app:estimators_m} for details on the learnable $\sM$. 
However, as depicted in the figure, the alternative estimator does not perform as effectively as ours.

\subsection{Classification of Dermatological Diseases}
\label{sec:semg}

\textbf{Dataset} 
Our problem is motivated by healthcare applications.
To show the effectiveness of our \ours{} method, we conduct experiments on the publicly available medical dataset known as Fitzpatrick 17k\footnote{Available at this
\href{https://github.com/mattgroh/fitzpatrick17k}{link}.}~\citep{groh2021evaluating}. This dataset provides two sets of labels for each patient's skin condition at varying levels of granularity.

At the highest level ($J=3$), skin conditions are categorized into three broad classes: benign lesions, malignant lesions, and non-neoplastic lesions. 
At a more detailed level ($K=9$), images are  classified into nine distinct categories.
We distribute the training dataset to the server and $C=10$ clients in an IID fashion. 
The percentage of samples on the server ($r$) relative to all training samples varies from $5\%$, $10\%$, to $20\%$.
Additional details are given in Appendix~\ref{sec:detail_medical}.

\begin{table}[ht]
\centering
\caption{\small \textbf{Experiments on real dataset.} Comparison of the accuracy for 9-class (black) and $3$-class (blue) on the Fitzpatrick dataset with different proportion of the dataset stored on the server relative to the overall training set. The best method is in boldface.}
\label{tab:real_data}
\begin{tabular}{lcccccc}
\toprule
$r$& Single &  FedTrans & FedRep & SemiFL & \ours{} (E)\\ 
\midrule
5\% & 38.00&39.11\color{blue}{(70.22)}&46.00\color{blue}{(71.67)}&44.44&\textbf{52.00}\color{blue}{(70.18)}\\ 
10\% & 50.44&44.66\color{blue}{(69.33)}&49.11\color{blue}{(72.10)}&52.22& \textbf{56.00}\color{blue}{(71.91)} \\
20\% & 53.11&54.00\color{blue}{(68.22)}&54.89\color{blue}{(72.29)}&57.11&\textbf{58.22}\color{blue}{(73.06)}\\
\bottomrule
\end{tabular}
\end{table}
Our primary objective is to predict the skin condition of each patient at $K=9$, based on their highest-level label at $J=3$. 
This prediction is achieved using a small subset of 9-class data obtained from the server.
Our \ours{} is compared to baselines in the CIFAR experiment. 
Given the unavailability of $\sM$ in real data, we estimate it from the data, and the corresponding result is labeled as FedMT(E). 

\noindent
\textbf{Results} 
The results are given in Tab.~\ref{tab:real_data}. 
It is evident that our \ours{}(E) method consistently outperforms other approaches across various proportions $r$, highlighting the robustness of our method even when a small amount of data is available within the desired label space. 
The advantage of FedMT (E) is particularly pronounced for smaller values of $r$.
For the 9-class accuracy, indicated in black, our method exhibited improvement from the best competitor's $46\%$ to $52\%$ when $r=5\%$. Moreover, when predicting at $K=9$, the performance of all methods improves with an increase in $r$, representing the proportion of supervision from the desired label space.
However, FedTrans demonstrates poorer performance as $r$ increases, owing to the reduction in samples across clients. For the 3-class accuracy, marked in blue, our method demonstrated a significant improvement, especially at a 20\% dataset proportion with an accuracy of 73.06\%, compared to 72.29\% for the next best method, FedRep. In this case, for the 9-class task, our method also led the field, achieving an accuracy of 58.22\% compared to 57.11\% for FedTrans, its closest competitor.
In summary, our methods exhibit superior performance for the 9-class task while also demonstrating commendable predictive capability for the 3-class task as an additional benefit.
\section{Discussion \& Conclusion}
\label{sec:conclusion}
\textbf{Algorithm cost}
The communication protocol of our algorithm precisely mirrors that of SemiFL~\citet{diao2022semifl}. 
Consequently, our algorithm does not incur additional costs compared to FedAvg.

\textbf{Privacy}
Our proposed method only shares model weights across clients and the server. 
Hence, we offer privacy guarantees equivalent to standard FL algorithms, such as FedAvg. 
Techniques for privacy-preserving FedAvg can be directly applied to our method.

\textbf{Theory vs practice}
Our generalization bound analysis in Theorem~\ref{thm:generalization_error} with an estimated correspondence matrix uses a pre-estimated matrix. 
We iteratively update our estimator in each round, leading to potential changes in the empirical loss over time. 
However, as depicted in Fig.~\ref{fig:cifar_goodness_of_hatm}, the estimated correspondence matrix converges. 
Hence, our analysis remains applicable to the estimated correspondence matrix when it reaches convergence.
We have used~\eqref{eq:m_hat} to estimate the correspondence matrix, it is an interesting future work to investigate other advanced methods such as~\citet{yao2020dual,zhang2021learning,li2022estimating} for its estimation.

In conclusion, we introduce~\ours{}, a novel and versatile FL framework designed to address the under-explored challenges posed by mixed-type labels. 
By incorporating a projection matrix into the client losses,~\ours{} demonstrates adaptability and effectiveness in various FL settings. 
Notably, its minimal modifications to local losses allow for seamless integration with various FL strategies beyond FedAvg.
Theoretically, we provide a new generalization bound to illustrate the effectiveness of combining data with different labeling criteria.
Our extensive experiments on both benchmark and medical datasets underscore the superior performance of~\ours{} compared to alternative methods.

\bibliographystyle{apalike}
\bibliography{ref}

\newpage
\appendix
\section{Notation Table}
\label{sec:appendix_notation}

\begin{table}[ht]
\caption{Important notations used in the paper.}
\centering
\begin{tabular}{cl}
\toprule
Notations & Description \\ 
\hline \\[-1.8ex]
$\tilde{\gO}$ & ${\cal O}(h(n) \log^k n) = \tilde{{\cal O}}(h(n)) $\\[0.1ex] 
$d$ & input dimension \\[0.1ex] 
$[m]$ & the set $\{1,2,\ldots, m\}$ where $m$ is a positive integer \\[0.1ex] 
$n$ & number of per desired label space class\\ 
&samples on the server\\[0.1ex]
$t$ & local updating iterations \\[0.1ex]
$C$ & number of clients in the FL, we index client by $c\in[C]$\\[0.1ex]
$J$ & number of classes in the other label space\\[0.1ex]
$K$ & number of classes in the desired label space\\[0.1ex]
$\sM$ & correspondence matrix\\[0.1ex]
$R$ & aggregation communication rounds \\[0.1ex]
$N_c$ & size of samples on the $c$-th client \\[0.1ex]
$S_{c}$ & the set of indices of observations on $c$-th client\\[0.1ex]
$\gX$ & input space $\gX\subset\R^d$ \\[0.1ex] 
$\tilde\gY$ & the other label space $\tilde\gY=\{\tilde Y^1,\tilde Y^2,\ldots,\tilde Y^J\}$\\[0.1ex] 
$\gY$ & desired label space $\gY=\{Y^1,Y^2,\ldots,Y^K\}$\\[0.1ex] 
$\gD_c$ & dataset $\mathcal{D}_c =\{(\vx_{i}^c, y_{i}^c): i\in S_{c}\}$ on $c$-th client \\[0.1ex]
$\gD_{s}$ & dataset on server $\gD_{s}=\{(\vx_i^s, y_i^s): i\in [nK]\}$ where $y_i^s\in \gY_{\text{fine}}$\\[0.1ex]
$\vx$ & inputs variable $\vx \in \mathcal{X}$\\[0.1ex]
$\vf(\cdot)$ &  a classification model that assigns a score to\\ 
&each of the $K$ classes for a given input $\vx$ \\[0.1ex]
$\ell(\cdot,\cdot)$ & loss function for the classification task \\[0.1ex]
$\hat R_{c}^{p}(\cdot)$ & empirical risk on $c$-th client \\[0.1ex]
$\hat R_{s}(\cdot)$ & empirical risk on the server \\[0.1ex]
$\hat R_{c}^{\text{reg}}(\cdot)$ & strong data augmentation loss regularized empirical risk on $c$-th client \\[0.1ex]
$\eta_{\text{local}}$ & local GD step size \\[0.1ex]
$\eta_{\text{agg}}$ & aggregation step size \\[0.1ex]
$\Delta_{K-1}$  & $K-1$-dimensional simplex \\
&$\Delta_{K-1} = \{\boldsymbol\pi: \sum_{k=1}^{K}\pi_k=1, \pi_k\in[0,1]\}$\\[0.1ex]
\bottomrule
\end{tabular}
\label{tab:notation}
\end{table}
\section{Technical Details of the generalization error}
\label{sec:appendix_theory}
In this section, we explain the technical details of analyzing the generalization error for our method. 
We introduce the notation in Section~\ref{app:notation_setting}, present all the technical lemmas in Section~\ref{app:lemma}, and provide the proof of the main result in Section~\ref{app:main_result}.

\subsection{Notation and problem setting}
\label{app:notation_setting}
In this section, we outline the notation and setting of the generalization error analysis for our method.
Consider a scenario where $(\vx, y, \tilde y) \in \gX\times \gY\times \tilde\gY$, with both $y$ and $\tilde y$ representing labels for the feature $\vx$ under two distinct labeling criteria. 
We operate under the constraint of observing only one label for each dataset. 
Let $\gD_{s} = {(\vx_i^s, y_i^s): i=1,\ldots, n}$ and $\gD_{c} = {(\vx_i^c, \tilde y_i^c): i=1,\ldots, m}$ be two datasets, and our objective is to learn a classifier $\vg: \gX \to \Delta_{K-1}$ based on the combined dataset $\gD_{s} \cup \gD_{c}$.

Define $\vf: \gX \to \Delta_{K-1}$ with its $k$th element as
\[f_k(\vx) = \sP(y = Y^{k} | X=\vx)\]
and $\sM: \Delta_{K-1}\to \Delta_{J-1}$ where
$\sM_{jk} = \sP(\tilde{y} = \tilde{Y}^{j}|y = Y^{k})$.

Also, define $\tilde{\vf}: \gX \to \Delta_{J-1}$ with its $j$th element as
\[\tilde{f}_j(\vx) = \sP(\tilde y = \tilde{Y}^{j} | X=\vx)\] 
and $\sS: \Delta_{J-1}\to \Delta_{K-1}$ where $\sS_{kj} = \sP(y = Y^{k}|\tilde{y} = \tilde{Y}^{j})$.
Based on the total law of probability and the conditional independent assumption, we have
\[\sP(\tilde{y}= \tilde{Y}^{j}|y = Y^{k},\vx) = \sP(\tilde{y}= \tilde{Y}^{j}|y = Y^{k})\quad\text{and}\quad \sP(y = Y^{k}|\tilde{y}= \tilde{Y}^{j},\vx) = \sP(y = Y^{k}|\tilde{y}= \tilde{Y}^{j}):\]
This leads to the expressions:
\begin{equation*}
    \begin{split}
        f_{k}(\vx) = \sP(y = Y^{k} | X=\vx) =& \sum_{j} \sP(y = Y^{k},\tilde{y} = \tilde{Y}^{j}| X=\vx)\\
        =&\sum_{j} \sP(y = Y^{k}|\tilde{y} = \tilde{Y}^{j}, \vx)\sP(\tilde{y} = \tilde{Y}^{j}|\vx)\\
        =&\sum_{j}\sP(y = Y^{k}|\tilde{y} = \tilde{Y}^{j})\sP(\tilde{y} = \tilde{Y}^{j}|\vx) = [\sS \tilde{\vf}(\vx)]_{k}.
    \end{split}
\end{equation*}
and
\begin{equation*}
    \begin{split}
     \tilde{f}_{j}(\vx) = \sP(\tilde{y} = \tilde{Y}^{j} | X=\vx) =& \sum_{k} \sP(y = Y^{k},\tilde{y} = \tilde{Y}^{j}| X=\vx)\\
        =&\sum_{k} \sP(\tilde{y} = \tilde{Y}^{j} | y = Y^{k},\vx)\sP(y = Y^{k}|\vx)\\
=&\sum_{k}\sP(\tilde{y} = \tilde{Y}^{j} | y = Y^{k})\sP(y = Y^{k}|\vx) = [\sM \vf(\vx)]_{j}.
    \end{split}
\end{equation*}
Viewing both $\vf$ and $\tilde\vf$ as functionals, we have the relationship:
\begin{equation}
\label{eq:two_function_correspondence}
\tilde\vf = \sM\vf\quad\text{and}\quad\vf = \sS \tilde\vf.
\end{equation}
Additionally, we find that:
\begin{equation*}
    \sS_{kj} = \sP(y = Y^{k}|\tilde{y} = \tilde{Y}^{j}) = \frac{\sP(\tilde{y} = \tilde{Y}^{j}|y = Y^{k})\sP(y = Y^{k})}{\sP(\tilde{y} = \tilde{Y}^{j})} = \sM_{jk}\frac{\sP(y = Y^{k})}{\sP(\tilde{y} = \tilde{Y}^{j})} = \frac{J}{K}\sM_{jk}
\end{equation*}
where the last equality is based on the assumption that both $y$ and $\tilde{y}$ have uniform priors.
Recognizing the relationship between $\sM$ and $\sS$, we may interchangeably refer to $\sM$ as known (estimated) when we actually mean $\sS$ is known (estimated), eliminating the necessity of introducing additional notations.

\subsection{Technical lemmas}
\label{app:lemma}
\begin{lemma}[Hoeffding's inequality]
\label{lemma:hoeffding}
Let $X_1,\ldots, X_n$ be IID random variables with $a_i\leq X_i \leq b_i$ for all $i$, then for any $\epsilon >0$,
\[\sP(|\bar{X} - \sE\bar{X}|\geq \epsilon) \leq 2\exp\left(-\frac{2n^2\epsilon^2}{\sum_{i=1}^{n}(b_i-a_i)^2}\right)\]
where $\bar{X} = (X_1+\ldots + X_n)/n$.
\end{lemma}
\begin{lemma}
\label{lemma:difference_bound}
Let $\vf$ and $\vg$ are two mappings from $\gX \to \Delta_{K-1}$, let $\{\vx_i^{s}\}_{i=1}^{n}$ are IID samples in $\gX$.
We then have 
\[\sE\|\vg(\vx) - \vf(\vx)\|_2 \leq \frac{1}{n}\sum_{i=1}^{n} \|\vg(\vx_i^s) - \vf(\vx_i^s)\|_2 + \sqrt{\frac{K\log n}{4n}}\]
with a probability of at least $1-2n^{-1/2}$.
\end{lemma}

\begin{proof}[Proof of Lemma~\ref{lemma:difference_bound}]
Note that we have $\|\vg(\vx_i^s) - \vf(\vx_i^s)\|_2\leq \sqrt{K}$ is bounded.
By Hoeffding's inequality in Lemma~\ref{lemma:hoeffding}, we have
\[\sP\left(|\sE\|\vg(\vx) - \vf(\vx)\|_2 -\frac{1}{n}\sum_{i=1}^{n} \|\vg(\vx_i^s) - \vf(\vx_i^s)\|_2 |\geq \epsilon \right) \leq 2\exp\left(-\frac{2n^2\epsilon^2}{\sum_{i=1}^{n}(b_i-a_i)^2}\right)\]
Let $\epsilon = \sqrt{K\log n/(4n)}$, and we obtain
\[\sE\|\vg(\vx) - \vf(\vx)\|_2 \leq \frac{1}{n}\sum_{i=1}^{n} \|\vg(\vx_i^s) - \vf(\vx_i^s)\|_2+ \sqrt{\frac{K\log n}{4n}} \]
with a probability of at least $1-2n^{-1/2}$.
\end{proof}

\begin{lemma}[Estimation error of $\hat\sS$]
\label{lemma:hatm_error}
Let the notation be as before.
Recall that $\sS_{kj} = \sP(y = Y^{k}|\tilde{y} = \tilde{Y}^{j})$ and 
\[\hat\sS_{kj} = \frac{m^{-1}\sum_{i=1}^{m} \mathbbm{1}(\tilde{y}_i^c = \tilde{Y}^{j}, y_i^c = Y^{k})}{m^{-1}\sum_{i=1}^{m} \mathbbm{1}(\tilde{y}_i^c = \tilde{Y}^{j})}.\]
The with probability of at least $1-2m^{-1/2}$, we have
\[\|\sS-\hat\sS\|_2 \leq C\sqrt{\frac{\log m}{m}}\]
for some constant $C>0$ that does not depend on $m$.
\end{lemma}

\begin{proof}[Proof of Lemma~\ref{lemma:hatm_error}]
Define 
\[\sC_{jk} = \sP(\tilde{y} = \tilde{Y}^{j}, y = Y^k) \quad\text{and}\quad \hat\sC_{jk} = m^{-1}\sum_{i=1}^{m}\mathbbm{1}(\tilde{y}_i^c = \tilde{Y}^{j}, y_i^c = Y^k)\]
and
\[D_{j} = \sP(\tilde y = \tilde{Y}^j) \quad\text{and}\quad \hat D_{j} = m^{-1}\sum_{i=1}^{m}\mathbbm{1}(\tilde{y}_i^c = Y^j).\]
Then
\begin{align*}
    \|\sS-\hat\sS\|_2^2 \leq & \|\sS-\hat\sS\|_F^2 =\sum_{j,k}(\hat\sS_{kj} -\sS_{kj})^2\\
    =&\sum_{j,k} \left(\frac{\hat\sC_{kj}}{\hat D_j} - \frac{\sC_{kj}}{D_j}\right)^2 = \sum_{j,k} \left[\hat \sC_{kj}(\hat D_j^{-1} - D_j^{-1}) + \frac{\hat \sC_{kj} - \sC_{kj}}{D_j}\right]^2\\
    \leq & 2\sum_{j,k} \left[\hat \sC_{kj}^2(\hat D_j^{-1} - D_j^{-1})^2 + \frac{(\hat \sC_{jk} - \sC_{kj})^2}{D_j^2}\right]\\
    \leq & 2\sum_{j} (\hat D_j^{-1} - D_j^{-1})^2 + 2J^2 \sum_{j,k} (\hat \sC_{kj} - \sC_{kj})^2.
\end{align*}
Note that $\mathbbm{1}(\tilde{y}_i^c = \tilde{Y}^{j}, y_i^c = Y^k) \overset{\text{ind}}{\sim} \text{Bern}(\sC_{jk})$, then by Bernstein's inequality, we have
\[\sP(|\hat{\sC}_{kj}-\sC_{kj}|\geq \epsilon )\leq 2\exp(-2m\epsilon^2). \]
Let $\epsilon = \sqrt{\log m/4m}$, then we have with a probability of at least $1-2m^{-1/2}$, $|\hat{\sC}_{kj}-\sC_{kj}|\leq \frac{\log m}{4m}$.
Similarly, we have that $\hat D_j > D_j - \frac{\log m} {4m})$ with a probability of at least $1-m^{-1/2}$.
Therefore, with high probability, we also have
\[|\hat D_j^{-1} - D_j^{-1}| \leq \frac{|\hat D_j - D_j|}{D_j(D_j + \log m/4m)}\leq C' \frac{\log m}{m}.\]
Therefore, we have
\[\|\sS-\hat\sS\|_2\leq C\sqrt{\frac{\log m}{m}}.\]
This completes the proof.
\end{proof}

\subsection{Generalization bound}
\label{app:main_result}
Let the notation be the same as before.
In this section, we show the prediction error in both label spaces.
Let us consider two cases. The first case is when $\sM$ is known, and the second case is when $\sM$ is estimated.

\noindent
\textbf{Case 1: $\sM$ is known.}
For prediction in $\gY$ when $\sM$ is known, we show that with a probability of at least $1-2\max(n,m)^{-1/2}$, we have
\begin{equation}
\label{eq:fine_bound_knownM}
\begin{split}
\sE\|\vg(\vx)-\vf(\vx)\|_2 \leq&  \underbrace{\frac{1}{m+n}\left\{\sum_{i\in [|\gD_s|]} \|\vg(\vx_i^s) - \vf(\vx_i^s) \|_2 
+ \sum_{i\in [|\gD_c|]} \|\vg(\vx_i^c) - \sS\tilde\vf(\vx_i^c) \|_2\right\}}_{\text{empirical loss}}\\
&+ \sqrt{\frac{K (m\log m + n\log n)}{4(m+n)^2}}     
\end{split}
\end{equation}

\noindent
\textbf{Case 2: $\sM$ is estimated.}
For prediction in $\gY$ with estimated $\sM$, we show that with a probability of at least $1-2\max(n,m)^{-1/2}$, we have
\begin{equation}
\label{eq:fine_bound_estimatedM}        
\begin{split}
\sE\|\vg(\vx)-\vf(\vx)\| \leq&  \underbrace{\frac{1}{m+n}\left\{\sum_{i\in [|\gD_s|]} \|\vg(\vx_i^s) - \vf(\vx_i^s) \|_2 
+ \sum_{i\in [|\gD_c|]} \|\vg(\vx_i^c) - \hat\sS\tilde\vf(\vx_i^c) \|_2\right\}}_{\text{empirical loss}}\\
&+ \sqrt{\frac{K (m\log m + n\log n)}{4(m+n)^2}} + C\sqrt{\frac{\log m}{m}}.
\end{split}
\end{equation}

For both cases for prediction in $\tilde\gY$, we have 
\begin{equation}
\label{eq:coarse_bound}
\sE\|\sM\vg(\vx)-\tilde\vf(\vx)\|_2 \leq  \sqrt{K} \sE\|\vg(\vx)-\vf(\vx)\|_2.
\end{equation}

\begin{proof}[Proof of~\eqref{eq:coarse_bound}]
\begin{equation*}
\sE\|\sM\vg(\vx)-\tilde\vf(\vx)\|_2 = \sE\|\sM(\vg(\vx) - \vf(\vx))\| _2\leq \|M\|_2\sE\|\vg(\vx) - \vf(\vx)\|_2
\end{equation*}
where
\begin{equation*}
    \|M\|_2^2 = \lambda_{\max}(M^{\top} M) \leq \text{tr}(M^{\top} M) \leq K
\end{equation*}
and the proof completes.
\end{proof}

\begin{proof}[Proof of~\eqref{eq:fine_bound_knownM}]
    \begin{align*}
\sE\|\vg(\vx)-\vf(\vx)\|_2 =& \sE\left\|\frac{n}{m+n}(\vg(\vx) - \vf(\vx)) + \frac{m}{m+n}(\vg(\vx) - \sS \tilde\vf(\vx))\right\|_2 \\
\leq & \frac{n}{m+n}\underbrace{\sE\|\vg(\vx) - \vf(\vx)\|_2}_{T_1} + \frac{m}{m+n}\underbrace{\sE\|\vg(\vx) - \sS\tilde\vf(\vx)\|_2}_{T_2}
\end{align*}
where the first equality is due to~\eqref{eq:two_function_correspondence}.
We now bound $T_1$ and $T_2$ respectively.

For $T_1$, apply Lemma~\ref{lemma:difference_bound}, we have that with probability at least $1-2n^{-1/2}$
\begin{equation}
\label{eq:bound_T1}
\begin{split}
    T_1 = & \frac{1}{n}\sum_{i=1}^{n} \|\vg(\vx_i^s) - \vf(\vx_i^s) \|_2 + \left\{\sE\|\vg(\vx) - \vf(\vx)\|_2-\frac{1}{n}\sum_{i=1}^{n} \|\vg(\vx_i^s) - \vf(\vx_i^s)\|_2\right\} \\
    \leq & \frac{1}{n}\sum_{i=1}^{n} \|\vg(\vx_i^s) - \vf(\vx_i^s) \|_2 + \sqrt{\frac{K\log n}{4n}}.     
\end{split}
\end{equation}

For $T_2$, by~\eqref{eq:two_function_correspondence} and apply Lemma~\ref{lemma:difference_bound}, we have that with probability at least $1-2m^{-1/2}$
\begin{equation}
    \label{eq:bound_T2}
\begin{split}
    T_2 = & \frac{1}{m}\sum_{i=1}^{m} \|\vg(\vx_i^c) - \sS\tilde{\vf}(\vx_i^c) \|_2 + \left\{\sE\|\vg(\vx) - \sS\tilde{\vf}(\vx)\|_2-\frac{1}{m}\sum_{i=1}^{m} \|\vg(\vx_i^c) - \sS\tilde{\vf}(\vx_i^c) \|_2\right\} \\
    \leq & \frac{1}{m}\sum_{i=1}^{m} \|\vg(\vx_i^c) - \sS\tilde{\vf}(\vx_i^c) \|_2 + \sqrt{\frac{K\log m}{4m}}. 
\end{split}
\end{equation}

Combine~\eqref{eq:bound_T1} and~\eqref{eq:bound_T2}, we then get with probability at least $1-2\max(n,m)^{-1/2}$
\begin{equation*}
    \sE\|\vg(\vx)-\vf(\vx)\|_2 \leq  \underbrace{\frac{1}{m+n}\left\{\sum_{i\in [|\gD_s|]} \|\vg(\vx_i^s) - \vf(\vx_i^s) \|_2 + \sum_{i\in [|\gD_c|]} \|\vg(\vx_i^c) - \sS\tilde{\vf}(\vx_i^c) \|_2\right\}}_{\text{empirical loss}}+ \underbrace{\sqrt{\frac{K (m\log m + n\log n)}{4(m+n)^2}}}_{\text{goes to $0$}} 
\end{equation*}
which completes the proof.
\end{proof}

\begin{proof}[Proof of~\eqref{eq:fine_bound_estimatedM}]
Recall that
\[\sE\|\vg(\vx)-\vf(\vx)\|_2 \leq \frac{n}{m+n}\underbrace{\sE\|\vg(\vx) - \vf(\vx)\|_2}_{T_1} + \frac{m}{m+n}\underbrace{\sE\|\vg(\vx) - \sS\tilde\vf(\vx)\|_2}_{T_2}.\]
The bound for $T_1$ is the same as before.
For $T_2$, we have
\begin{equation}
\label{eq:bound_T2_estimatedM}
\begin{split}
T_2 =& \frac{1}{m}\sum_{i=1}^{m} \|\vg(\vx_i^c) - \hat\sS\tilde\vf(\vx_i^c) \|_2 + \left\{\sE\|\vg(\vx) - \sS\tilde\vf(\vx)\|_2-\frac{1}{m}\sum_{i=1}^{m} \|\vg(\vx_i^c) - \sS\tilde\vf(\vx_i^c) \|_2\right\} \\
&+ \frac{1}{m}\left\{\sum_{i=1}^{m}\|\vg(\vx_i^c) - \sS\tilde\vf(\vx_i^c) \|_2 - \|\vg(\vx_i^c) - \hat\sS\tilde\vf(\vx_i^c)\|_2\right\}\\
\overset{(a)}{\leq} & \frac{1}{m}\sum_{i=1}^{m} \|\vg(\vx_i^c) - \hat\sS\tilde\vf(\vx_i^c)\|_2 + \sqrt{\frac{K\log m}{4m}} + \frac{1}{m}\sum_{i=1}^{m}\|(\sS-\hat\sS)\tilde\vf(x_i^c)\|_2\\
\overset{(b)}{\leq}  & \frac{1}{m}\sum_{i=1}^{m} \|\vg(\vx_i^c) - \hat\sS\tilde\vf(\vx_i^c)\|_2 + \sqrt{\frac{K\log m}{4m}} + \|\sS-\hat\sS\|_2 \\
\leq  & \frac{1}{m}\sum_{i=1}^{m} \|\vg(\vx_i^c) - \hat\sS\tilde\vf(\vx_i^c)\|_2 + \sqrt{\frac{K\log m}{4m}} + C\sqrt{\frac{\log m}{m}}
\end{split}
\end{equation}
where $(a)$ is due to triangle inequality, $(b)$ is because $\tilde\vf$ is bounded by $1$, and the last inequality is based on Lemma~\ref{lemma:hatm_error}.

Combine~\eqref{eq:bound_T1} and~\eqref{eq:bound_T2}, we then get with probability at least $1-2\max(n,m)^{-1/2}$
\begin{equation*}
\begin{split}
\sE\|\vg(\vx)-\vf(\vx)\| \leq &  \underbrace{\frac{1}{m+n}\left\{\sum_{i\in [|\gD_s|]} \|\vg(\vx_i^s) - \vf(\vx_i^s) \| + \sum_{i\in [|\gD_c|]} \|\vg(\vx_i^c) - \hat\sS\tilde{\vf}(\vx_i^c) \|\right\}}_{\text{empirical loss}}+ \sqrt{\frac{K (m\log m + n\log n)}{2(m+n)^2}} \\
& + C \sqrt{\frac{\log m}{m}}.
\end{split}
\end{equation*}
which completes the proof.

\end{proof}

\section{Experimental Details}
\label{sec:appendix_exp}
We present our model architectures and training details of the benchmark and the publicly available Fitzpatrick 17k clinical dataset in this section.

\subsection{Model Architecture and Implementation Details on Benchmark}
\label{sec:detail_benchmark}
The details of the experiments on the benchmark CIFAR100 dataset are presented in this section. 

\noindent
\textbf{Model architecture}
For our benchmark experiments on CIFAR100, we use ResNet18 as our backbone unless otherwise specified.
The network details are listed in Tab.~\ref{tab:model_cifar100}.
We also used MobileNet-V2 and ResNet50 as different backbones for comparison.
We do not repeat the network architectures of MobileNet-V2 and ResNet50 in this section.
We use the model architecture from the \texttt{torchvision} module in Pytorch 1.8.1.
For both ResNet18 and ResNet50, we skip the first maxpooling layer.

\begin{table}[ht]
\caption{Model architecture of the benchmark experiment on CIFAR100. For convolutional layer (Conv2D), we list parameters with sequence of input and output dimension, kernel size, stride and padding. For max pooling layer (MaxPool2D), we list kernel and stride. For fully connected layer (FC), we list input and output dimension. For BatchNormalization layer (BN), we list the channel dimension.}
\label{tab:model_cifar100}
\renewcommand\arraystretch{1.2}
\centering
\scalebox{0.99}{
\begin{tabular}{l|cccc}
\toprule
\multicolumn{1}{c|}{\multirow{1}{*}{Layer}} & \multicolumn{4}{c}{Details} \\ \hline
\multicolumn{1}{c|}{\multirow{1}{*}{1}}& \multicolumn{4}{c}{Conv2D(3, 64, 7, 2, 3), BN(64), ReLU} \\ \hline
\multicolumn{1}{c|}{\multirow{1}{*}{2}}& \multicolumn{4}{c}{Conv2D(64, 64, 3, 1, 1), BN(64), ReLU} \\ \hline
\multicolumn{1}{c|}{\multirow{1}{*}{3}}& \multicolumn{4}{c}{Conv2D(64, 64, 3, 1, 1), BN(64)} \\ \hline
\multicolumn{1}{c|}{\multirow{1}{*}{4}}& \multicolumn{4}{c}{Conv2D(64, 64, 3, 1, 1), BN(64), ReLU} \\ \hline
\multicolumn{1}{c|}{\multirow{1}{*}{5}}& \multicolumn{4}{c}{Conv2D(64, 64, 3, 1, 1), BN(64)} \\ \hline
\multicolumn{1}{c|}{\multirow{1}{*}{6}}& \multicolumn{4}{c}{Conv2D(64, 128, 3, 2, 1), BN(128), ReLU} \\ \hline
\multicolumn{1}{c|}{\multirow{1}{*}{7}}& \multicolumn{4}{c}{Conv2D(128, 128, 3, 1, 1), BN(64)} \\ \hline
\multicolumn{1}{c|}{\multirow{1}{*}{8}}& \multicolumn{4}{c}{Conv2D(64, 128, 1, 2, 0), BN(128)} \\ \hline

\multicolumn{1}{c|}{\multirow{1}{*}{9}}& \multicolumn{4}{c}{Conv2D(128, 128, 3, 1, 1), BN(128), ReLU} \\ \hline
\multicolumn{1}{c|}{\multirow{1}{*}{10}}& \multicolumn{4}{c}{Conv2D(128, 128, 3, 1, 1), BN(64)} \\ \hline

\multicolumn{1}{c|}{\multirow{1}{*}{11}}& \multicolumn{4}{c}{Conv2D(128, 256, 3, 2, 1), BN(128), ReLU} \\ \hline
\multicolumn{1}{c|}{\multirow{1}{*}{12}}& \multicolumn{4}{c}{Conv2D(256, 256, 3, 1, 1), BN(64)} \\ \hline

\multicolumn{1}{c|}{\multirow{1}{*}{13}}& \multicolumn{4}{c}{Conv2D(128, 256, 1, 2, 0), BN(128)} \\ \hline

\multicolumn{1}{c|}{\multirow{1}{*}{14}}& \multicolumn{4}{c}{Conv2D(256, 256, 3, 1, 1), BN(128), ReLU} \\ \hline
\multicolumn{1}{c|}{\multirow{1}{*}{15}}& \multicolumn{4}{c}{Conv2D(256, 256, 3, 1, 1), BN(64)} \\ \hline

\multicolumn{1}{c|}{\multirow{1}{*}{16}}& \multicolumn{4}{c}{Conv2D(256, 512, 3, 2, 1), BN(512), ReLU} \\ \hline
\multicolumn{1}{c|}{\multirow{1}{*}{17}}& \multicolumn{4}{c}{Conv2D(512, 512, 3, 1, 1), BN(512)} \\ \hline

\multicolumn{1}{c|}{\multirow{1}{*}{18}}& \multicolumn{4}{c}{Conv2D(256, 512, 1, 2, 0), BN(512)} \\ \hline

\multicolumn{1}{c|}{\multirow{1}{*}{19}}& \multicolumn{4}{c}{Conv2D(512, 512, 3, 1, 1), BN(512), ReLU} \\ \hline
\multicolumn{1}{c|}{\multirow{1}{*}{20}}& \multicolumn{4}{c}{Conv2D(512, 512, 3, 1, 1), BN(512)} \\ \hline
\multicolumn{1}{c|}{\multirow{1}{*}{21}}& \multicolumn{4}{c}{AvgPool2D}\\ \hline
\multicolumn{1}{c|}{\multirow{1}{*}{22}}& \multicolumn{4}{c}{FC(512, 100)} \\ \bottomrule
\end{tabular}%
}
\end{table}

\subsubsection{Details of the data splitting}
\label{sec:non-iid-split}
We describe how to prepare the CIFAR100 training dataset into sub-class data on the server and super-class datasets on the clients. 
We first select $n$ samples from each of the $K=100$ sub-classes as the datasets on the server. 
Then for the rest of the observations in the CIFAR100 training set, we split then into $C$ pieces each with $N_c$ observations and each client is assigned $N_c$ samples.

\noindent
\textbf{IID split}
The $N_c$ samples are assigned to clients completely at random and are balanced in subclasses.
For example, when $N_c=4000$, each client has $400$ observations from each of the $100$ sub-classes.
Hence in this case $\sM$ is a $20\times 100$ matrix, its $(j,i)$-th element is $1$ if the $i$-th class is in $j$-th superclass and $0$ otherwise.

\noindent
\textbf{Non-IID split} 
We follow the non-iid splitting in~\citet{lu2021federated} to create data heterogeneity in the benchmark dataset CIFAR100.
Specifically, we partition the dataset among clients using the following criteria.
\begin{enumerate}
    \item We initiate the non-IID setting by assigning a random prior to each class, where the prior probability is proportional to a uniform variable ranging from [0.35, 0.55].

    \item Subsequently, we systematically select $2$ majority classes on each client, with the prior probability for these two classes set to $4/(1-\gamma)$. 
    The value of $\gamma$ is tuned to three different levels: $0.1$, $0.9$, and $0.99$. A higher $\gamma$ signifies a greater degree of heterogeneity.
    Following class selection, we normalize the prior probabilities to ensure their summation equals one.

    \item Finally, we sample $N_c=4000$ instances on each client of various classes according to these probabilities.
\end{enumerate}

\noindent
\textbf{Training details}
We implement all the methods by PyTorch 1.8.1 and conducted all the experiments on an NVIDIA Tesla V100 GPU. 
We consider two different initialization schemes on CIFAR100, namely random initialization and initialization with the weights pretrained on ImageNet.
For the weights pretrained on ImageNet, we use the default pretrained model in \texttt{torchvision}.
We set the local update step to be $1$, that is each model updates one epoch then aggregates with the others.

We first state the parameter setting for \ours{} with random initialization.
We use the SGD optimizer~\citep{ruder2016overview} with weight decay $5\times 10^{-4}$ and momentum 0.9 and learning rate $0.03$ with a cosine decay scheduler. The number of clients is set to 10.
The local training iteration on clients is 5 and total communication iterations is set to be 200. The relative weight of mix loss to fix loss is 1. The Mixup hyperparameter a is set to be 0.75.
For additional results in the Appendix~\ref{sec:appendix_more}, we specify the hyper-parameters only when they are different.

We describe details of the baseline methods and corresponding hyper-parameters as follows.
\begin{itemize}
   
    \item \textbf{Single} For this baseline method, we only use the limited data on the server to train a $K$-class classifiers with the objective is the classical CE loss
    \begin{equation*}
    \hat R_{s}(\vf;\gD^{s}) = -\frac{1}{nK}\sum_{i=1}^{nK} \mathbbm{1}(y_i^s= Y^k)\log \sigma(f_{k}(\vx_i))
    \end{equation*}
    
    \item \textbf{FedRep} \citep{collins2021exploiting}  For this baseline method, all clients train a super-class classifier and the server trains a sub-class classifier that differs in the last layer with different output dimension.   
    Let $\vf$ be the backbone and $\mW_{Y}$ and $\mW_{\tilde Y}$ are matrices of appropriate dimensions that map to $\Delta_{K}$ and $\Delta_{J}$ respectively.
    The learning objective is
    
    \begin{equation*}
    \widehat{R}(\vf, \mW_{Y}, \mW_{\tilde Y})=\sum_{c=1}^{C}\hat R_{c}(\mW_{Y}\vf;\gD^{c})+\hat R_{s}(\mW_{\tilde Y}\vf;\gD^{s}),
    \end{equation*}
     where 
    \begin{equation*}
    \hat R_{c}(\vf;\gD^{c}) = -\frac{1}{N_c}\sum_{i\in S_{c}} \sum_{j=1}^{J}\mathbbm{1}(y_i=\tilde Y^j)\log \sigma(f_{j}(\vx_i))
    \end{equation*}
    and $\hat R_{s}(\vf;\gD^{s})$ is the same as the objective function in the Single baseline approach.
    All classifiers on the server and on the clients use the same ResNet18 backbone, but they have different last layers.
    For the classifiers on clients, their last layer is FC(512,20); for the classifier on the server, its last layer is FC(512,100).
    The last layers of classifiers on clients are also not shared during the training.
    The rest of the parameters of all classifiers are aggregated using FedAvg during training. 
    We use the same optimizer as our \ours.\footnote{Note that the data labeling settings, models, and evaluation methods implemented in our work is adjusted to make fair comparison with other methods and different from those in \citet{collins2021exploiting}.}
    
    \item \textbf{FedTrans} 
    For this baseline method, we first pretrain a model using the client data to train a $J$-class classifiers using FedAvg with probability projection or label projection.
    This pretrained net is then fine-tuned on the server.
    During the fine-tuning, we add a new linear layer with $K$ classes with random initialization and backbone is initialized with the pretrained model.
    We use the same optimizer as our \ours{} during the pretrain step and the model is fine-tuned on the server for $100$ epochs with SGD optimizer and learning rate $0.01$.

    \item \textbf{SemiFL} The method in~\citet{diao2022semifl}. We use the official implementation on \href{https://github.com/diaoenmao/SemiFL-Semi-Supervised-Federated-Learning-for-Unlabeled-Clients-with-Alternate-Training}{Gitub} in our experiment for comparison. We set the loss weight of mix loss relative to fix loss to 1. In most cases, we set the threshold for selecting reliable pseudo label to 0.95. 

\end{itemize}

\subsection{Model Architecture and Implementation Details on Medical Dataset}
\label{sec:detail_medical}

\noindent
\textbf{Data preprocessing}
Due to the inaccessibility of certain images in the Fitzpatrick 17k dataset, we have filtered out these unavailable images.
At the highest level ($J=3$), skin conditions are categorized into three broad classes: $2231$ benign lesions, $2260$ malignant lesions, and $12035$ non-neoplastic lesions. 
For a more detailed analysis ($K=9$), images are further classified into nine distinct categories: $10841$ labeled as inflammatory, $1352$ as malignant epidermal, $1194$ as genodermatoses, $1067$ as benign dermal, $931$ as benign epidermal, $570$ as malignant melanoma, $233$ as benign melanocyte, $182$ as malignant cutaneous lymphoma, and $156$ as malignant dermal.
As indicated by the summary statistics, the Fitzpatrick 17k dataset exhibits an extreme class imbalance, with over $65\%$ of the samples belonging to the ``inflammatory" class, while the ``malignant dermal" class comprises only around $0.9\%$ of the samples. 
This significant imbalance poses a challenge, potentially causing skewed model learning where the algorithm leans towards favoring the majority class, thereby compromising performance on minority classes. 
To address this issue, we implement data augmentation techniques to mitigate the extremeness of the training dataset.

To ensure a fair comparison, we initially reserve a test dataset for evaluating the performance of all approaches. This test dataset is created by randomly selecting $50$ images from each of the $K=9$ classes, resulting in a total of $450$ images. 
Subsequently, the remaining dataset is processed to construct the training dataset.
We begin by randomly sampling $5,000$ samples from the ``inflammatory" class. For the remaining classes, if the number of samples in the training set is below $200$, we augment the class to $1000$. If the count is between $200$ and $500$, we augment it to $1200$; if it falls between $500$ and $1000$, we augment to $1400$. Otherwise, we augment to $1600$.
This strategy results in a more balanced dataset, mitigating biases towards majority classes, and enhancing the model's ability to generalize and accurately predict minority classes.

\noindent
\textbf{Training details}
We employ ResNet18 with ImageNet pretrained model weights as initialization across all experiments. 
We first state hyperparameter setting for FedMT.
The learning rete is 0.01 with cosine decay scheduler. And we also use SGD as optimizer with weight decay $5\times 10^{-4}$ and momentum 0.9. The overall iteration is set to be 100 and local training iteration on each client is 2. For other baseline methods, we use the same setting except local training set to be 1.
For both SemiFL and our method, we set the pseudo-label threshold value $\rho$ to $0.95$.
\section{More Experimental Results on CIFAR100}
\label{sec:appendix_more}
In this section, we present additional experimental findings based on the CIFAR100 benchmark dataset. 
In Section~\ref{app:initializations}, we outline the test accuracy obtained through two distinct initialization strategies: random initialization and the utilization of pre-trained weights from ImageNet.
Moving to Section~\ref{sec:coarse_prediction}, we provide experiment details on using our trained model to do prediction in the other label space.
Furthermore, Section~\ref{sec:hyper-parameters} offers supplementary results that illustrate the performance of various approaches under different hyper-parameter configurations.

\subsection{Different initialization strategies}
\label{app:initializations}
We consider two different initialization strategies for our experiment.
The first strategy is that all model weights are initialized randomly, and the second strategy is that all model weights are initialized using ImageNet pretrained model weights.
We present the result of two different initialization strategies for the baseline approaches in Tab.~\ref{tab:cifar_initialization}.
The ImageNet pretrained weight initialization leads to better results for all baselines.
\begin{table*}[htbp]
\centering
\caption{\small \textbf{Baselines using different initialization strategies.} Comparison of the accuracy for 100-class classification of baseline methods on the CIFAR100 benchmark dataset with different per sub-class number $n$ images on the server using different initialization strategies. We report mean{\tiny{$\pm$sd}} from three trial runs.}
\label{tab:cifar_initialization}
\begin{tabular}{lccc|ccc}
\toprule
\multirow{2}{*}{$n$} & \multicolumn{3}{c|}{Random initialization} &  \multicolumn{3}{c}{ImageNet pretrained weight initialization}\\
\cmidrule{2-7}
& Single &  FedRep & FedTrans  & Single &  FedRep & FedTrans \\ 
\midrule
5 & 8.56\scriptsize{$\pm$0.33} & 16.89\scriptsize{$\pm$0.74} & 21.31\scriptsize{$\pm$1.27} & 15.80\scriptsize{$\pm$0.66} & 39.45\scriptsize{$\pm$0.49} & 43.43\scriptsize{$\pm$0.67} \\
10 & 12.29\scriptsize{$\pm$0.32}  & 21.90\scriptsize{$\pm$1.27} &  24.53\scriptsize{$\pm$1.40} & 20.38\scriptsize{$\pm$0.28}  & 42.97\scriptsize{$\pm$0.15} & 48.70\scriptsize{$\pm$0.29} \\
15& 14.34\scriptsize{$\pm$0.15}  & 23.39\scriptsize{$\pm$0.67} & 26.21\scriptsize{$\pm$0.95} & 27.16\scriptsize{$\pm$0.68} & 45.01\scriptsize{$\pm$0.26} & 50.49\scriptsize{$\pm$0.55} \\
20 & 16.23\scriptsize{$\pm$0.21}  & 24.70\scriptsize{$\pm$0.55} & 27.49\scriptsize{$\pm$0.85} & 29.12\scriptsize{$\pm$0.80} & 46.57\scriptsize{$\pm$0.19} & 53.14\scriptsize{$\pm$0.34} \\
25 & 17.73\scriptsize{$\pm$0.20} &  26.53\scriptsize{$\pm$0.29} & 28.16\scriptsize{$\pm$1.30} &  29.57\scriptsize{$\pm$0.29} & 47.37\scriptsize{$\pm$0.20} & 54.83\scriptsize{$\pm$0.30} \\
\bottomrule
\end{tabular}
\end{table*}

\subsection{Details of coarse label prediction}
\label{sec:coarse_prediction}
In this section, we provide the details pertaining to Figure~\ref{fig:cifar_coarse}. 

For the FedTrans approach, we employ the $J$-class model directly trained on client data using FedAvg as the model for coarse label prediction. 
Consequently, this approach remains uninfluenced by any server data, resulting in a constant prediction accuracy across various values of $n$.

In the case of the FedRep approach, predictions on the test dataset are made using each of the $J$-class classifiers on the client side, and the average test accuracy across clients is reported in the figure.

As for our \ours{} approach, we use our finalized trained model to predict the $K$-class scores. 
Subsequently, we project these scores to the $J$-class using either the true correspondence matrix $\sM$ in FedMT (T) or the estimated matrix $\hat\sM$ in FedMT (E) for making predictions. 
It is worth noting that the estimated $\hat\sM$ is recorded during training, eliminating the need for additional effort in computing the matrix.

\subsection{Hyper-parameters in \ours{} method}
\label{sec:hyper-parameters}

\noindent
\textbf{Threshold of pseudo label}
The threshold $\gamma$ plays a crucial role as a hyper-parameter in our method, influencing the construction of high-confidence samples. 
Therefore, we present the test accuracy of \ours{} (E) alongside the baseline SemiFL for various choices of $\gamma$. 
The table highlights that for both approaches, within the confines of a fixed computational budget (\ie, a fixed number of communication rounds), opting for a lower value of $\gamma$ results in improved performance.
It is noticeable that performance may potentially improve for higher values of $\gamma$ if we increase communication rounds. 
However, such an enhancement inevitably comes at the cost of increased computational resources.

\begin{table}[htbp]
\centering
\caption{\small Comparison of the accuracy for 100-class classification with different threshold for selecting pseudo labels on the CIFAR100 benchmark dataset with $n=25$ images per-subclass on the server. The best method is highlighted in boldface.}
\label{tab:cifar_threshold}
\begin{tabular}{lccccc}
\toprule
 Threshold &0.7 &  0.8 & 0.9 &  0.95 \\ 
\midrule
SemiFL  & 58.48&58.71&56.00&55.13 \\
\midrule
{\textit{Ours}}: FedMT (E) & \bf 61.39&\bf 61.00&\bf 59.71& \bf 56.76\\
\bottomrule
\end{tabular}
\end{table}

\noindent
\textbf{Penalty weight}
Another critical hyper-parameter in our methodology is the weight $\lambda_2$ in the client loss~\eqref{eq:strong_augmented_local_loss}. In the cases of both FedMT (T) and FedMT (E), the test accuracy demonstrates an initial rise with increasing values of $\lambda_2$.
However, after reaching $1/16$, the test accuracy begins to decrease as the penalty strength increases.
Consequently, we tune this weight in our experiments based on this observed pattern.

\begin{table}[htbp]
\centering
\caption{\small Comparison of the accuracy for 100-class classification with different loss weight on the CIFAR100 benchmark dataset with $n=25$ images per-subclass on the server. The best hyper-parameter is highlighted in boldface.}
\label{tab:cifar_loss_weight}
\begin{tabular}{lcccccccc}
\toprule
$\lambda_2$ & 1 & 1/2 & 1/4 & 1/8 & 1/16 & 1/32 & 1/64 \\ 
\midrule
{\textit{Ours}}: FedMT (T) & 58.63& 60.82& 64.02 & 66.28 & \bf 66.90 & 66.28&  66.29\\
{\textit{Ours}}: FedMT (E)& 56.76& 56.87& 60.18& 63.89& \bf 65.93 & 65.06& 64.76\\
\bottomrule
\end{tabular}
\end{table}

\subsection{Different estimators for $\sM$}
\label{app:estimators_m}
We explore two strategies for estimating the correspondence matrix. 
The first approach involves calculating the matrix through~\eqref{eq:m_hat}, while the second approach treats the matrix as a learnable parameter during the training process.

In the case of estimating the correspondence matrix via~\eqref{eq:m_hat}, each client has its client-specific estimator for the matrix, which is used to construct the client loss function~\eqref{eq:strong_augmented_local_loss} in the FedMT (E) approach. 
For the FedMT (E) + Average approach, we first estimate the correspondence matrix on each client. 
Subsequently, these estimators are aggregated through averaging, and the aggregated matrix is then transmitted back to each client for computing the loss~\eqref{eq:strong_augmented_local_loss}.

When treating the correspondence matrix as a learnable parameter, a numerical trick is used to ensure that it remains a well-defined correspondence matrix, adhering to the conditions $\sM_{jk}\in [0,1]$ for all $j\in[J]$, $k\in[K]$, and $\sum_{j} \sM_{jk} = 1$ for all $k\in[K]$. Specifically, by introducing free learnable parameters $m_{jk}\in \mathbb{R}$, we set 
\[\sM_{jk} = \frac{\exp(m_{jk})}{\sum_{j'}\exp(m_{j'k})},\]
and the unconstrained optimization is performed over ${m_{jk}}$ which are initialized randomly. 
In this approach, the learnable correspondence matrix is allowed to differ for each client.
For the learnable + average approach, we average the learned correspondence matrix from the previous round across clients and utilize this matrix as the initial value for the current round.

We demonstrate the results of the four methods in a non-iid setting in Tab.~\ref{tab:cifar_estimator}.
\begin{table}[htbp]
\centering
\caption{\small \textbf{ Performance for different estimation methods under non-IID setting.}  Comparison of the accuracy for 100-class classification with estimation methods for correspondence matrix on the CIFAR100 benchmark dataset with $n=25$ images per-subclass on the server. The best method is highlighted in boldface.}
\label{tab:cifar_estimator}
\begin{tabular}{lcccc}
\toprule
 & IID & \multicolumn{3}{c}{Non-IID}\\ 
& &0.1 & 0.9 & 0.99\\
\midrule

{\textit{Ours}}: FedMT (E) & \bf{66.01}& \bf{65.47}& \bf{58.64}& \bf{41.09}\\
{\textit{Ours}}: FedMT (E) + Average & 63.14& 62.50 & 48.33& 33.92\\
\hline
Learnable&56.51& 55.25& 49.49& 36.55\\
Learnable + Average & 56.03& 55.61 & 49.45& 37.09\\
\bottomrule
\end{tabular}
\end{table}
The table clearly indicates that in both IID and non-IID settings, our present FedMT (E) exhibits the most favorable performance. 
The experimental outcomes for the Learnable and Learnable + Average approaches serve as a sanity check, as each local client minimizes the local loss. 
In contrast, the average, being non-optimal for each client, inevitably results in a performance degradation.

\end{document}